\documentclass[journal,12pt,onecolumn,draftclsnofoot,]{IEEEtran}

\usepackage{import}
\usepackage{amsmath,amssymb}
\usepackage{mathtools}
\mathtoolsset{showonlyrefs,showmanualtags}
\usepackage{enumitem}
\usepackage{cancel}
\usepackage{graphicx}
\usepackage{bbm}
\usepackage[normalem]{ulem}
\usepackage{hyperref}
\usepackage{amsthm}
\usepackage[ruled]{algorithm2e}
\usepackage{multirow,array}

\usepackage{siunitx}

\usepackage{standalone}
\usepackage{pgfplots}
\usepackage{tikz}
\usepackage{subcaption}

\usepackage{import}
\usepackage{amsmath,amssymb}
\usepackage{mathtools}
\usepackage{etoolbox}

\makeatletter
\let\pragma@iinput=\@iinput
\def\@iinput#1{\xdef\@pragmafile{#1}\pragma@iinput{#1}}
\def\@pragmafile{default}
\def\pragmaonce{%
	\csname pragma@\@pragmafile\endcsname
	\global\expandafter\let \csname pragma@\@pragmafile\endcsname =  
}
\makeatother

\DeclareMathOperator*{\argmin}{arg\,min}

\ifdefined\ceil
\else
\DeclarePairedDelimiter\ceil{\lceil}{\rceil}

\fi
\DeclarePairedDelimiterX{\inprd}[2]{\langle}{\rangle}{#1, #2}
\newcommand{\inprdS}[2]{\left\langle#1,#2\right\rangle}

\newcommand{\ellnrm}[1]{\ell_{#1}}
\newcommand{\LnrmS}[1]{\left\lVert #1 \right\rVert}
\newcommand{\Lnrm}[1]{\lVert #1 \rVert}

\newcommand{\EV}{{\mathbb{E}}}
\newcommand{\EVS}[1]{{\mathbb{E}\left[#1\right]}}

\newcommand{\Rel}{\mathbb{R}}

\newcommand{\bigO}{\mathcal{O}}

\newcommand{\allzeros}{\mathbf{0}}

\newcommand{\cc}[1]{\mathcal{#1}}

\newcommand{\calD}{\cc{D}}

\definecolor{darkgreen}{rgb}{0.0, 0.5, 0.0}

\newcommand{\stkout}[1]{\ifmmode\text{\sout{\ensuremath{#1}}}\else\sout{#1}\fi}

\newcommand{\fig}[1]{{Fig.~\ref{fig:#1}}}
\newcommand{\tbl}[1]{{Table~\ref{tbl:#1}}}
\newcommand{\secn}[1]{{Sec.~\ref{secn:#1}}}
\newcommand{\apdx}[1]{{App.~\ref{secn:#1}}}

\newcommand{\thrm}[1]{{Theorem~\ref{thrm:#1}}}
\newcommand{\colly}[1]{{Corollary~\ref{colly:#1}}}

\newcommand{\algo}[1]{{Alg.~\ref{algo:#1}}}
\newcommand{\eqn}[1]{{\eqref{eqn:#1}}}

\makeatletter
\newcommand*\ifcounter[1]{%
	\ifcsname c@#1\endcsname
	\expandafter\@firstoftwo
	\else
	\expandafter\@secondoftwo
	\fi
}
\makeatother

\ifcsmacro{theorem}{}{

	\ifcounter{chapter}{
		
	}{
		
	}
	
	\newtheorem{theorem}{Theorem}
	\newtheorem{corollary}{Corollary}
	
	\newcounter{example}[section]
	
	\newcounter{problem}[section]
	
}

\def\enumtheoremstart{\begin{enumerate}[noitemsep,label=(\roman*)]}
	\def\enumtheoremend{\end{enumerate}}

\usepackage{transparent}

\newif\ifshowcomments
\newif\ifshowdeleted

\newcommand{\devnull}[1]{}

\newcommand{\printcomment}[2]{\incolor{#1}{[[\ifx#1\empty\else#1: \fi#2]]}}

\newcommand{\ifequals}[3]{\ifthenelse{\equal{#1}{#2}}{#3}{}}
\newcommand{\case}[2]{#1 #2} 
\newenvironment{switch}[1]{\renewcommand{\case}{\ifequals{#1}}}{}

\definecolor{darkred}{rgb}{0.8, 0.01, 0.1}
\definecolor{darkgreen}{rgb}{0.0, 0.5, 0.0}
\definecolor{darkorange}{rgb}{0.93, 0.35, 0.1}

\newcommand{\incolor}[2]{\ignorespaces
	\begin{switch}{#1}\ignorespaces
		\case{TA}{\color{darkred}}\ignorespaces
		\case{SD}{\color{darkorange}}\ignorespaces
		\case{SK}{\color{darkgreen}}\ignorespaces
		\case{}{\color{red}}\ignorespaces
		#2
	\end{switch}
}

\usepackage{cite}
\usepackage{etoolbox}

\begin{document}
\title{
Compressing gradients by exploiting temporal correlation in momentum-SGD
}

\author{Tharindu~B.~Adikari 
and~Stark~C.~Draper
\thanks{
This work was supported in part by Huawei Technologies Canada; and in part by the Natural Science and Engineering Research Council (NSERC) of Canada through a Discovery Research Grant. 
(\textit{Corresponding author: Tharindu~B.~Adikari.})
}
\thanks{
This paper was presented in part at the 11th International Symposium on Topics in Coding (ISTC), Montreal, QC, Canada, August 2021, and the paper has been accepted for publication in the IEEE Journal on Selected Areas in Information Theory (JSAIT) Volume 2, Issue 3 (2021), \url{https://ieeexplore.ieee.org/document/9511618}.
}
\thanks{
The authors are with the Electrical and Computer Engineering
Department, University of Toronto, Toronto, ON M5S 2E4, Canada 
(e-mail: tharindu.adikari@mail.utoronto.ca; stark.draper@utoronto.ca).
}
}

%
%

%

\maketitle

\begin{abstract}
An increasing bottleneck in decentralized optimization is communication. Bigger models and growing datasets mean that decentralization of computation is important and that the amount of information exchanged is quickly growing. While compression techniques have been introduced to cope with the latter, none has considered leveraging the temporal correlations that exist in consecutive vector updates. An important example is distributed momentum-SGD where temporal correlation is enhanced by the low-pass-filtering effect of applying momentum. In this paper we design and analyze compression methods that exploit temporal correlation in systems both with and without error-feedback. Experiments with the ImageNet dataset demonstrate that our proposed methods offer significant reduction in the rate of communication at only a negligible increase in computation complexity. We further analyze the convergence of SGD when compression is applied with error-feedback. In the literature, convergence guarantees are developed only for compressors that provide error-bounds point-wise, i.e., for each input to the compressor. In contrast, many important codes (e.g. rate-distortion codes) provide error-bounds only in expectation and thus provide a more general guarantee. In this paper we prove the convergence of SGD under an expected error assumption by establishing a bound for the minimum gradient norm.
\end{abstract}


%


\pragmaonce  

\newcommand{\loss}{f}
\newcommand{\gloss}{g}
\newcommand{\vvw}{{w}}
\newcommand{\momentum}{v}
\newcommand{\hX}{\hat{X}}
\newcommand{\hY}{\hat{Y}}
\newcommand{\young}{\xi}
\newcommand{\stepsize}{\eta}
\newcommand{\numDists}{{n}}
\newcommand{\dimw}{{d}}
\newcommand{\fQ}{Q}
\newcommand{\tbeta}{\tilde{\beta}}

\newcommand{\wi}[2][]{\vvw^{#1}_{#2}}
\newcommand{\twi}[2][]{\tilde{\vvw}^{#1}_{#2}}
\newcommand{\gi}[2][]{\gloss^{#1}_{#2}}
\newcommand{\ei}[2][]{e^{#1}_{#2}}
\newcommand{\ri}[2][]{r^{#1}_{#2}}
\newcommand{\rhi}[2][]{\hat{r}^{#1}_{#2}}
\newcommand{\rti}[2][]{\tilde{r}^{#1}_{#2}}
\newcommand{\vi}[2][]{\momentum^{#1}_{#2}}
\newcommand{\vti}[2][]{\tilde{\momentum}^{#1}_{#2}}
\newcommand{\vhi}[2][]{p^{#1}_{#2}}
\newcommand{\ui}[2][]{u^{#1}_{#2}}
\newcommand{\uti}[2][]{\tilde{u}^{#1}_{#2}}
\newcommand{\taui}[2][]{\tau^{#1}_{#2}}
\newcommand{\gti}[2][]{\tilde{\nabla}^{#1}_{#2}}
\newcommand{\gdi}[2][]{\nabla^{#1}_{#2}}

\newcommand{\setriI}{\{\ui[i]{t}[k] \mid k\in\ccIit\}}
\newcommand{\ccIitp}[2][]{\mathcal{J}^{#1}_{#2}}
\newcommand{\ccIit}{\ccIitp[i]{t}}
\newcommand{\sumworkers}{{\sum_{i=1}^{\numDists}}}
\newcommand{\avgopn}{\frac{1}{\numDists}\sum_{i\in[\numDists]}}
\newcommand{\fQT}{{\fQ_{\text{Top-$K$}}}}

\newcommand{\fPln}{P_{\text{Lin}}}
\newcommand{\fPek}{P_{\text{Est-$K$}}}

\section{Introduction}
Decentralized optimization has become the norm for training machine
learning models on large datasets.  With the need to train bigger
models on ever-growing datasets, scalability has become a key focus in
the research community.  While an obvious solution to growing dataset
size is to increase the number of workers, communication amongst
workers can turn into the processing bottleneck.  For popular
benchmark models such as AlexNet, ResNet and BERT, communication time
can account for a significant portion of the overall training
time \cite{alistarh2017qsgd, seide20141, lin2018deep}.  The BERT
(``Bidirectional Encoder Representing from Transformers") architecture
for language models \cite{devlin2018bert} comprises about 340 million
parameters.  If $32$-bit floating-point representation is used,
transmission of one gradient update amounts to around 1.3GB
($340 \times 10^6$ parameters $\times$ 32 bits per parameter $\times$
$2^{-33}$ gigabytes per bit $\approx$ 1.3GB).  Frequent communication
of such large payloads can easily overwhelm a network, resulting in
greatly prolonged training time.  As another example, in certain types
of decentralized systems, such as federated learning, communication
may be inherently limited since federated learning employs mobile
devices as worker nodes.  For these reasons communication remains an
important bottleneck in decentralized optimization and the development
of methods that reduce communication overhead, while not (too greatly)
impacting convergence rate, is of utmost importance.

One solution is to apply compression to the gradient (or more
generally to the messages exchanged between workers and the master) to
alleviate the communication bottleneck.  There has been an expanding
literature on gradient compression within the last few years,
e.g.,~\cite{seide20141, alistarh2017qsgd, wen2017terngrad,
bernstein2018signsgd, wu2018error, karimireddy2019error,
zheng2019communication}.  Compression schemes have been demonstrated
to work well with distributed stochastic gradient descent (SGD) and
its variants.  For example in \cite{bernstein2018signsgd}
and \cite{zheng2019communication} the authors consider both SGD and
SGD with momentum (momentum-SGD).  In the latter, compression is
applied to the momentum vector instead of the gradient vector.
However, these methods fail to exploit the memory, or temporal correlations, that
exists in the stream of update vectors. In this paper we
introduce novel compression schemes to leverage such temporal
correlations.

Possible sources of correlations are many-fold.  In gradient descent
(when gradients are exact) for smooth objective functions we expect
the gradients to be close to each other.  Typically the smoothness of
a function is described by the $L$-Lipschitz gradients assumption.
When gradients are not exact, as in the case of SGD, the stochasticity
of gradients introduces noise that may partially conceal correlations
across gradients.  Such concealment may be remedied by computing
gradients using larger minibatches. 
Our focus is on momentum-SGD which is a better candidate for temporal correlations. 
The consecutive update vectors in momentum-SGD have minimal changes, yielding high correlations. 

One classic and widely applied approach for compressing sources with memory is predictive coding. 
In more recent work, predictive coding has been used for compression of datasets~\cite{barowsky2021predictive}. 
Motivated by differential pulse code modulation (DPCM)~\cite{wiegand2011source}, in this paper we consider a predictive coding based approach for the compression of momentum-SGD updates. 
We investigate systems both with and without error-feedback. 
We demonstrate that, when quantization is
combined with a simple predictor in presence of temporally correlated vectors, 
a significant reduction in the
rate of communication is realized.  In the case of error-feedback, the
application of the same predictor (i.e., the same predictor used in non-error-feedback systems) does not realize a reduction in the communication rate. 
We therefore develop an algorithm that estimates and extrapolates the correlated momentum vectors. 
We note that our proposed methods are not using very advanced algorithms for prediction. 
While our methods demonstrate the usefulness of exploiting temporal correlation with predictive coding, more advanced ideas should certainly perform even better.
In \secn{contrib} we outline our contributions in detail.

\subsection{Related work}
The most common gradient compression schemes achieve compression by sparsifying the gradient vector or by lowering the precision of its components. The
methods based on sparsification such as
Top-$K$~\cite{strom2015scalable, dryden2016communication,
aji2017sparse, lin2017deep, lin2018deep},
Rand-$K$ \cite{wangni2018gradient, stich2018sparsified} and
Spectral-ATOMO \cite{wang2018atomo} preserve only the most significant
gradient elements, effectively reducing the quantity of information
(the number of gradient components communicated).  For example,
Top-$K$ compression only preserves the $K$ components of the gradient
that are largest in magnitude (as well as their positions in the
gradient vector).  On the other hand, methods based on quantization
such as 1-bit SGD~\cite{seide20141}, {QSGD}~\cite{alistarh2017qsgd},
TernGrad~\cite{wen2017terngrad}, ZipML~\cite{zhang2017zipml} and
{SignSGD}~\cite{bernstein2018signsgd} reduce the overall
floating-point precision of the gradient.  Sign-based scalar
quantizers such as Scaled-sign, SignSGD and
Signum~\cite{bernstein2018signsgd} sit at the far end of
quantization-based algorithms.  Such schemes quantize real values to
only two levels, $\pm a$ for some scalar $a$.  When quantization is
restricted to 1-bit scalar quantization, and the source distribution
is uni-modal and symmetric, then this (scalar) quantization approach
is optimal.  And, indeed, it has been empirically demonstrated that in
models such as ResNet-20 and ResNet-50 the distribution of the
components in the gradient is uni-modal and
symmetric~\cite{glorot2010understanding, bernstein2018signsgd,
shi2019understanding}, motivating the use of these symmetric 1-bit
quantizers.

The authors of \cite{yu2018gradiveq} propose a method that exploits joint statistics between different components of the gradient vector, i.e., spatial correlations.
The authors observe that the joint statistics change only slowly through the iterations (specifically, the covariance matrix), allowing the same estimate of the covariance matrix to be reused in their compression algorithm for a number of iterations. 
In other words, the joint statistics are near-stationary. 
In contrast, the methods we discuss in this paper exploit different time samples of the same component in the gradient vector, i.e., the temporal correlations amongst iterates.

Although gradient compression yields large communication savings,
quantization errors due to compression affects convergence rates.
This has been demonstrated both empirically and theoretically. 
{The authors of~\cite{karimireddy2019error} theoretically show that SGD with Scaled-sign 
compression may not converge by providing an example of non-convergence.} 
Inspired by concept of noise shaping in Delta-Sigma modulation~\cite{oppenheim2001discrete}, the authors
of~\cite{seide20141} employ an error-feedback mechanism to improve the
empirical convergence of SGD with compression.  Error-feedback, also
referred to as error-compensation, computes the quantization error and
add it to the input of the quantizer in the next iteration.  This way
the error gets transmitted eventually, albeit with a delay.  For
smooth functions for which the gradient does not change considerably
over a small number of iterations, the delay {can be shown not
significantly to} impact the rate of convergence.  Convergence
guarantees have been developed for error-feedback schemes when
the compression scheme employed is a
``$\delta$-compressor''~\cite{stich2018sparsified,
karimireddy2019error, zheng2019communication}, defined as follows.
Let $X\in\Rel^\dimw$ and $\hX\in\Rel^\dimw$ where the later is the
reconstruction of $X$ after compression.  A compression scheme is
known as a $\delta$-compressor if for any input $X$,
$\Lnrm{X-\hX}^2\leq(1-\delta)\Lnrm{X}^2$ for some $\delta\in(0,1]$.
Scaled-sign and Top-$K$ are $\frac{1}{\dimw}$-approximate and
$\frac{K}{\dimw}$-approximate compressors respectively.  
As we will discuss further below, schemes
such as Top-$K$ and Rand-$K$ offer nearly the same performance as SGD
while yielding very high compression rates when employed in
conjunction with error-feedback.

\subsection{Contributions} \label{secn:contrib}
Our contributions are both algorithmic and analytic.  First, we
introduce a novel approach that exploits the temporal
correlations that exist in momentum-SGD update vectors. 
Momentum can be understood
as the application of an exponentially weighted low-pass filter (LPF) to gradients across
iterations.  This essentially filters out high-frequency components
and only preserves low-frequency ones. This reduces the variation in
the resulting update vectors in consecutive iterations, making them
good candidates for the application of predictive coding. 
The key idea behind predictive coding is simple. 
Instead of inputting the momentum vector to the quantizer, compute a prediction of the momentum vector, compute the error in the prediction, and then input the vector of prediction errors to the quantizer. 
Strong temporal correlations in momentum vector lead to an accurate prediction. 
For such predictions, the variance of the components in the prediction error vector is lower than those in the original momentum vector. 
Therefore, the prediction error vector can be encoded using a lower bit rate. 
In~\secn{withoutef} we first consider systems that
do not employ error-feedback and introduce our baseline algorithm for prediction. 
In this setting we demonstrate that employing a simple linear predictor in combination with Scaled-sign and Top-$K$ quantizers can offer significant rate savings over their counterparts without prediction.

Next, in~\secn{algorithmef} we turn to systems that
operate \emph{with} error-feedback. These systems require a more
careful treatment as the employment of error-feedback affects 
the statistics of the components in the compressed vector. 
We show that the rate savings realized by the previous (linear) predictor 
quickly vanish when error-feedback is employed.
In this section we consider momentum-SGD in conjunction with Top-$K$ as the baseline.
We introduce the novel prediction scheme we name Est-$K$ (``Estimated-Top-$K$'') that works in conjunction with the Top-$K$ quantizer. 
Est-$K$ incorporates a momentum estimator that extrapolates the momentum update from past observations for each of the undescribed components in the update vector (those outside the Top-$K$). In our tests we observe this
idea reduces the payload of Top-$K$ by about $40\%$. 
Overall, and to the best of our knowledge, this paper is the first presentation of the exploitation of temporally correlated updates in momentum-SGD for the purpose of compression. 

We evaluate the performance of the algorithms we consider by training a wide residual network classifier \cite{zagoruyko2016wide} using the down-sampled ImageNet dataset in~\cite{chrabaszcz2017downsampled}. 
\tbl{topklsummary} summarizes our experimental results with different quantizers. 
The table is divided into three sections: the baseline momentum-SGD algorithm (without compression), the algorithms that do not use error-feedback and the algorithms that do use error-feedback. 
The second column indicates whether the algorithms exploit temporal correlations through a prediction scheme. 
The predictor used in the second and third sections of the table are the linear predictor we discuss in~\secn{withoutef} and the Est-$K$ predictor we discuss in~\secn{algorithmef} respectively. 
The {top-5}~\footnote{Please note ``top-5'' is an evaluation metric distinct from the ``Top-$K$'' compression algorithm, an unfortunate similarity in naming.} test accuracy, i.e., whether the correct category is in the list of 5 produced by the algorithm, is presented.  We tweaked parameters in compression algorithms to keep the test accuracy roughly equal within each section. 
All algorithms use momentum parameter $\beta=0.99$.  The dimension of
the gradient is $\dimw$.  Final accuracy is obtained after 28 training
epochs.  The last column indicates the average number of 
bits used per gradient component in each iteration.  Final test
accuracy is top-$5$. Note that ``Top-$K$-Q'' is Top-$K$ with
quantization, as is discussed in \cite{dryden2016communication}.

\begin{figure}
\centering\includegraphics[width=0.5\textwidth]{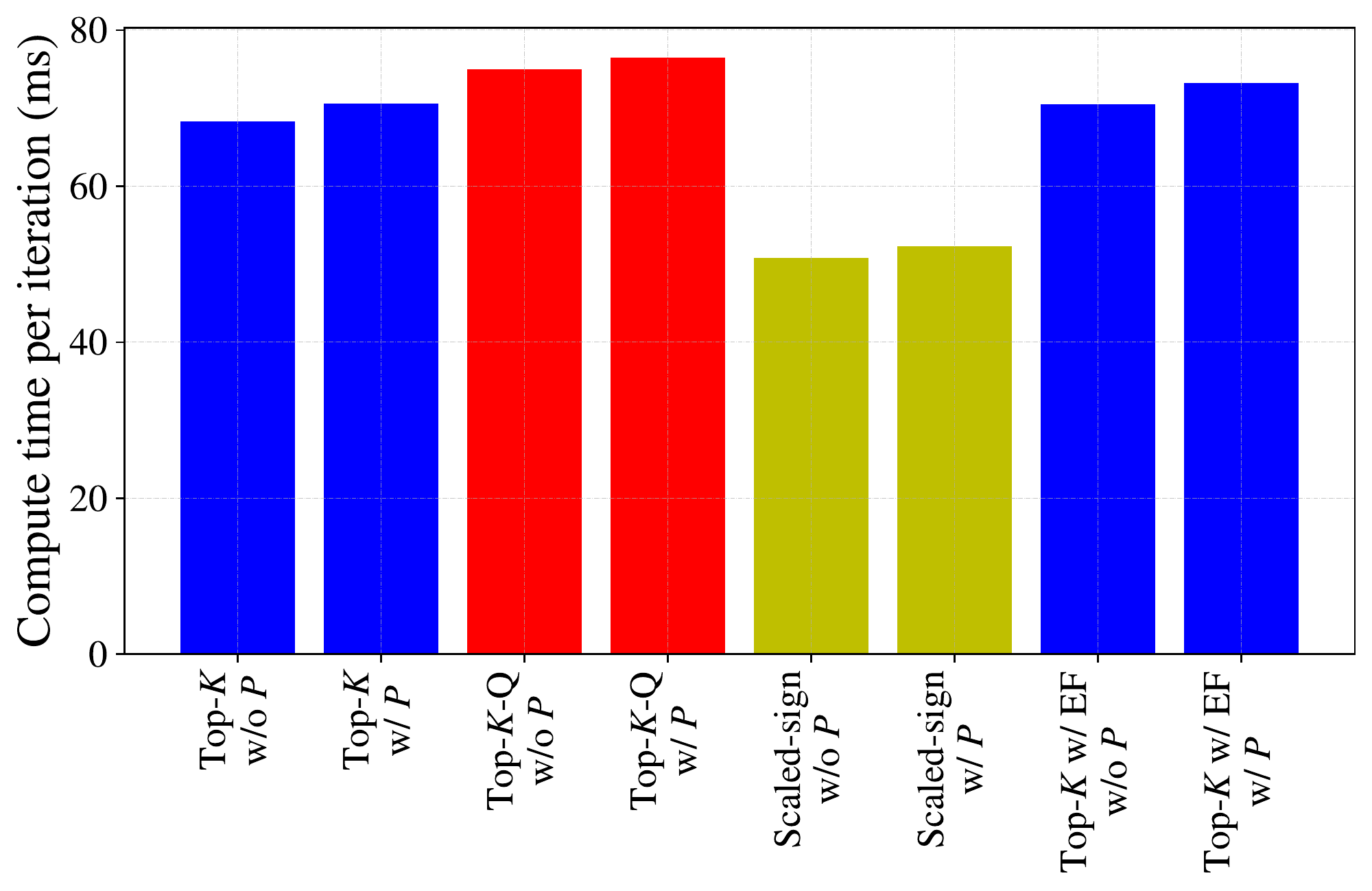}
\caption{Average computation time per iteration (excluding time for communication). 
Computations are gradient calculation, quantization, and prediction. 
The experiments that do and do not use prediction are denoted as w/ $P$ and w/o $P$, respectively. 
Only the right-most two experiments employ error-feedback.}
\label{fig:imagenet32_compute_time}
\end{figure}

In \tbl{topklsummary} the baseline is momentum-SGD without compression and it reaches $61.8\%$ test accuracy. 
Although this value is below the state-of-the-art for ImageNet, the reduction is to be expected as we use a down-sampled version of ImageNet, therefore, comprising of less informative images. 
Authors of \cite{chrabaszcz2017downsampled} provide a baseline for this (down-sampled) dataset wherein the final test accuracy varies between $60\%$ and $70\%$ for different learning rates (see Fig.~2 therein). 
Note that $K$ is a parameter that controls the compression rate in Top-$K$ and Top-$K$-Q. 
The main take-away of the table is the following. 
In either category (with or without error-feedback) the algorithms that use a predictor that exploits temporal correlation incur less communication overhead for the same performance than does the same algorithm without a predictor. This demonstrates the usefulness of exploiting temporal correlations.
In \fig{imagenet32_compute_time} we present a comparison of the average per iteration computation time of a worker, for each of the experiments with compression that are tabulated in \tbl{topklsummary}. 
In \fig{imagenet32_compute_time} we observe that for each quantizer the time per iteration when using prediction is only slightly higher than when not using prediction.
This added time serves as a measure of the very small additional computational complexity required workers to implement prediction.

\begin{table*}
\renewcommand{\arraystretch}{1.4}
\centering
\begin{tabular}{|c|c|cccc|}
\hline
\multicolumn{2}{|c|}{\textbf{Compressor}}
& \begin{tabular}{@{}c@{}}\textbf{Employs} 				\\ \textbf{error-feedback}	\end{tabular}
& \begin{tabular}{@{}c@{}}\textbf{Exploits temporal} 	\\ \textbf{correlations} 	\end{tabular}
& \begin{tabular}{@{}c@{}}\textbf{Final test} 			\\ \textbf{accuracy \%}		\end{tabular}
& \begin{tabular}{@{}c@{}}\textbf{Bits per} 			\\ \textbf{component} 		\end{tabular} 					\\
\hline\hline
\multicolumn{2}{|c|}{Baseline (no compression)}             	& --    & --            &  61.8  	& 32 			\\
\hline\hline
\multirow{2}{*}{Top-$K$}        &  $K=0.35\dimw$   				& No    & No            &  58.5  	& 12 			\\
\cline{2-6}
								&  $K=0.015\dimw$				& No    & \textbf{Yes}  &  58.5  	& 0.6 	        \\
\hline
\multirow{2}{*}{Top-$K$-Q}      &  $K=0.23\dimw$				& No    & No            &  58.5  	& 1.0 			\\
\cline{2-6}
								&  $K=0.01\dimw$				& No    & \textbf{Yes}  &  58.5  	& 0.1 			\\
\hline
\multirow{2}{*}{Scaled-sign}    &  --							& No    & No            &  58.5  	& 1.0 			\\
\cline{2-6}
								&  --							& No    & \textbf{Yes}  &  61.1  	& 1.0 			\\
\hline\hline
\multirow{2}{*}{Top-$K$} 		& $K=1.2\times10^{-4}\dimw$     & Yes   & No      		&  59.0 	& 0.0056 		\\
\cline{2-6}
								& $K=6.5\times10^{-5}\dimw$     & Yes  	& \textbf{Yes}  &  59.0 	& 0.0031 		\\
\hline
\end{tabular}
\caption{Summary of performance of the algorithms we consider.}
\label{tbl:topklsummary}
\end{table*}

On the analytic side, we contribute the the literature that analyzes
the convergence of SGD (without momentum) when error-feedback is
employed and compression is performed.  The literature contains
convergence guarantees for such systems when the compressors are
$\delta$-compressors.  However, the point-wise guarantees of
$\delta$-compressors are not satisfied by a wide range of interesting
compression algorithms.  Therefore, in this paper we study compression
guarantees that are only in expectation, with possible application of a prediction system. 
Our analysis very significantly expands the types of compression systems that can be employed with
guarantees to include rate-distortion codes, e.g., those that satisfy
a mean-squared error distortion constraint $\EV[\Lnrm{X-\hX}^2]\leq D$.

\subsection{Outline}
An outline of the remainder of the paper is as follows.
In \secn{sysmodel} we introduce the system model with which we work.
The model we introduce is general enough to encompass the variety of
scenarios that we explore, though the full generality of the model is
not employed until the later sections.  
In \secn{withoutef} we consider systems that do not employ error-feedback and introduce our linear prediction scheme.
In \secn{algorithmef} we consider systems that employ error-feedback and introduce our Est-$K$ prediction scheme. 
In \secn{convergenceproposed} we present the convergence analysis of SGD with compressors that satisfy a mean-squared error distortion constraint. 
While we present the respective numerical results in \secn{withoutef} and \secn{algorithmef}, we defer till 
\secn{numericaliclrtemporal} to discuss the setup of the 
numerical experiments conducted, therein providing details on
datasets, computational units, batch size, etc. We conclude
in \secn{conclusions} with thoughts on future work.  Some background,
algorithmic details, and convergence analysis are deferred to the
appendices. 
In \apdx{reasoningmastermom} we detail the reasoning for some choices made in our system model. 
Finally, in \apdx{proofsoftheorems} we provide proofs of the analytical results.

\subsection{Notation} For a real valued vector $x$, $x[j]$ denotes the $j$th
entry of $x$, 
and $\Lnrm{x}$ denotes the $\ellnrm{2}$-norm. 
We use upper-case to denote random vectors, e.g., $X$.
For $N\in\mathbb{Z}^+$ we denote the index set $\{1,\dots,N\}$ by $[N]$. 
For Big-O notation we use $\bigO$, 
and $\allzeros$ denotes the all-zeros vector of dimension $\dimw$.


\section{System model} \label{secn:sysmodel}

\begin{figure*}
\centering\includegraphics[width=0.8\textwidth]{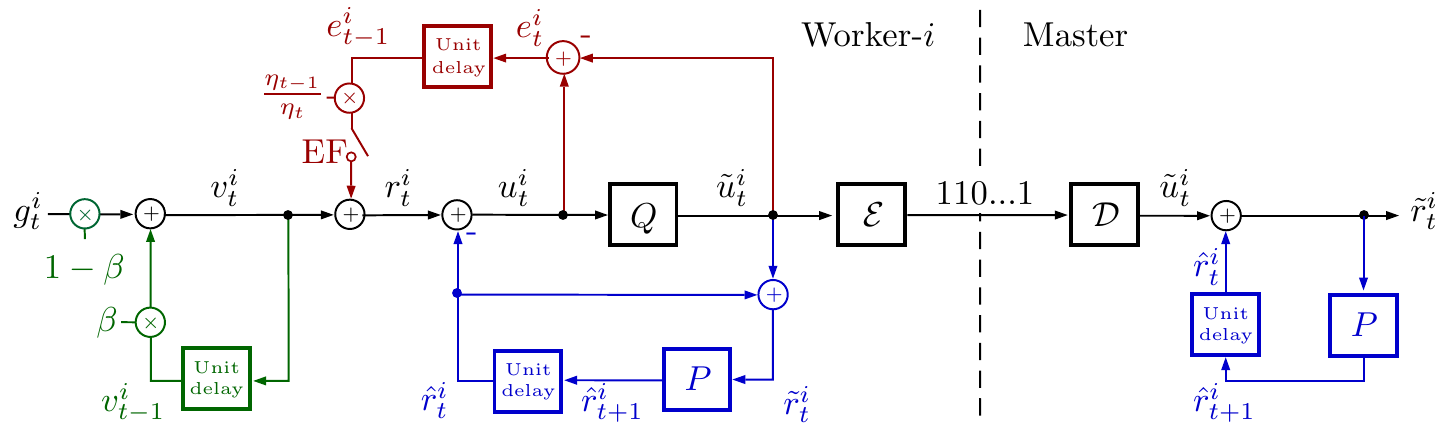}
\caption{System diagram for master and one worker (worker-$i$) that includes the components for momentum (green), prediction (blue), and error-feedback (red), along with the quantizer $\fQ$, encoder $\mathcal{E}$ and decoder $\mathcal{D}$.}
\label{fig:block_diagram}
\end{figure*}

Let $\loss:\Rel^\dimw\to\Rel$ be a differentiable function. 
We consider a master-worker system where $\numDists$
workers perform iterative optimization. 
The goal of the system is to calculate $w^\ast = \argmin_w f(w)$. 
As already discussed, we focus on momentum-SGD and the design of worker-to-master compression methods. The system model with which we work
is depicted in \fig{block_diagram}. We now formally describe the functioning of the system at the worker-side and at the master-side.

\subsection{Worker-side operations}
In iteration
$t\in\{0,\dots,T-1\}$ and for worker index $i\in[\numDists]$ the
system operates as follows.  All the following vectors are in
$\Rel^\dimw$.
\begin{subequations} \label{eqn:sysModmain}
\begin{align}
\vi[i]{t} &= \beta\vi[i]{t-1} + (1-\beta)\gi[i]{t} \label{eqn:momupdateeqn_sysMod} \\
\ri[i]{t} &= \vi[i]{t} + {\frac{\stepsize_{t-1}}{\stepsize_{t}}} \ei[i]{t-1} \label{eqn:errorfb_sysMod} \\
\ui[i]{t} & = \ri[i]{t}-\rhi[i]{t} \label{eqn:prederror_sysMod} \\
\uti[i]{t} & = \fQ(\ui[i]{t})\label{eqn:quant_sysMod} \\
\ei[i]{t} &= \ui[i]{t} - \uti[i]{t} \label{eqn:errorfbri_sysMod} \\
\rti[i]{t} & = \uti[i]{t} + \rhi[i]{t} \label{eqn:observation_sysMod} \\
\rhi[i]{t+1} & = P(\rti[i]{t}).
\label{eqn:errorfbQPredict_sysMod}
\end{align}
\end{subequations}
In iteration $t$ worker-$i$ computes the stochastic gradient
$\gi[i]{t}$ that satisfies the property
$\EV[\gi[i]{t}] = \nabla \loss(\wi{t})$ where $\wi{t}$ is the current estimate of the
optimum optimization variable.  When we discuss the operation of the
master below we will assume the system is synchronized so all workers
have the same $\wi{t}$. Eq.~\eqn{momupdateeqn_sysMod} is the heavy ball momentum update
performed at the worker with $\vi[i]{-1}=\allzeros$.  The momentum
parameter $0\leq\beta<1$ controls the bandwidth of the low-pass
filter.  In practice $\beta$ is set close to $1$, e.g. $0.9$ or
$0.99$.  Eq.~\eqn{errorfb_sysMod} is the result of the (possible)
error-feedback; note the presence of the switch labeled ``EF'' which
can be either open or closed.  The initial error vector $\ei[i]{-1}$
is set to all-zeros.  The learning rate is denoted by $\stepsize_{t}$
and is initialized to $\stepsize_{-1}=0$. 

Eq.~\eqn{prederror_sysMod} is the prediction error calculation, where $\rhi[i]{t}$ 
is the prediction of $\ri[i]{t}$ as per the prediction system we define later. 
Respectively, \eqn{quant_sysMod} and \eqn{errorfbri_sysMod} are the quantizer and the quantization error. 
Note the quantizer works on the entire vector $\ui[i]{t} \in \Rel^\dimw$. 
Eq.~\eqn{observation_sysMod} computes $\rti[i]{t}$, the reconstruction of $\ri[i]{t}$ after prediction and quantization. 
Finally, \eqn{errorfbQPredict_sysMod} is the prediction system which is available at both the worker-$i$ and the master. 
This system uses $\rti[i]{t}$ to compute $\rhi[i]{t+1}$, the prediction of $\ri[i]{t+1}$. 
In $t=0$ we initialize $\rhi[i]{0}$ to the all-zeros vector. The algorithms that we discuss in \secn{withoutef} and \secn{algorithmef} differ in how we design the predictor $P$. Note the predictor can have memory. For ease of reference, we tabulate system parameters in \tbl{sysparamstable}.

\begin{table}[]
	\renewcommand{\arraystretch}{1.4}
	\caption{List of system parameters.}
	\label{tbl:sysparamstable}
	\centering
	\begin{tabular}{|c|c|}
		\hline
		\textbf{Parameter}        & \textbf{Definition}    			          \\ \hline\hline
		$i$, $t$ & worker index, iteration index 		 	                  \\ \hline
		$\wi{t}$        & model parameter vector 				              \\ \hline
		$\gi[i]{t}$     & stochastic gradient 				 	              \\ \hline
		$\vi[i]{t}$     & momentum vector 				 	 	              \\ \hline
		$\ri[i]{t}$     & momentum with possible error-feedback	              \\ \hline
		$\rhi[i]{t}$ 	& prediction of $\ri[i]{t}$		                  	  \\ \hline
		$\ui[i]{t}$  	& quantizer input (prediction error for $\ri[i]{t}$)  \\ \hline
		$\uti[i]{t}$ 	& quantizer output    				                  \\ \hline
		$\ei[i]{t}$ 	& quantization error for $\ui[i]{t}$ 			      \\ \hline
		$\rti[i]{t}$ 	& reconstruction of $\ri[i]{t}$ 		              \\ \hline
\end{tabular}\end{table}

Encoder $\mathcal{E}$ encodes $\uti[i]{t}$ and produces the bit stream that is sent to the master. 
Note that the bit stream a worker sends to the master is unique to each worker. 
For this reason, the worker-to-master communication is more bandwidth intensive than is the 
master-to-worker communication that we discuss next. 

\subsection{Master-side operations}
We now turn to the master.  The decoder $\mathcal{D}$ produces
$\uti[i]{t}$, the master's reconstruction of $\ui[i]{t}$.  The prediction system $P$ employed at the master is
same as the one employed at the worker-$i$.  The master computes $\rti[i]{t}$ as per \eqn{observation_sysMod}, and the average of $\rti[i]{t}$ across all workers (all $i$).  Finally, it
broadcasts the average back to the workers.  All workers update the
parameter vector $\wi{t}\in\Rel^\dimw$ as
$\wi{t+1} = \wi{t} - \stepsize_{t}\avgopn\rti[i]{t}$. 
Note that broadcasting delivers the same payload to all workers. There exist algorithms (e.g., MPI\_Bcast in the message passing interface) that efficiently implement this broadcasting operation. 
Therefore, master-to-worker communication is much less problematic as a source of communication bottleneck. 
For this reason, in this paper we consider only the compression of messages from workers to the master. 

In our system model presented in \fig{block_diagram} we consider the application of momentum at the worker. One can also imagine a system where the worker quantizes the stochastic gradient (without momentum being applied) and communicates to the master (with or without error-feedback), after which the master applies momentum. This approach of applying momentum at the master is not desirable due to error accumulation in the momentum computed at the master. 
In \apdx{reasoningmastermom} we further detail this reasoning for systems both with and without error-feedback. 
For the rest of the paper we consider the model presented in \fig{block_diagram}.

\subsection{An example of the functioning of the system} \label{secn:exampletopk}

The following example demonstrates how different components in \fig{block_diagram} work together to implement momentum-SGD with the Top-$K$ quantizer. 
We choose Top-$K$ since we use it to set baseline performance in both \secn{withoutef} and \secn{algorithmef}. 
The Top-$K$ scheme does not consist of a prediction component. Therefore, we set $P$ to the zero function, i.e., $\rhi[i]{t+1}=P(\rti[i]{t})=\allzeros$. 
This choice of $P$ is equivalent the removal of the blue color components in \fig{block_diagram}. 
For Top-$K$ we denote the corresponding quantizer $Q$ as $\fQT$. 
This quantizer produces a sparse vector $\uti[i]{t}$ by setting all elements in
$\ui[i]{t}$ to zero except the $K$ elements with the largest magnitudes. 
If the non-zero values are further quantized,
this becomes the Top-$K$-Q scheme discussed in \cite{dryden2016communication}.

Let $\ccIit$ be the index set of $K$ the non-zero locations in $\uti[i]{t}$. 
The function of the encoder $\mathcal{E}$ is to losslessly compress $\ccIit$ and the set of non-zero values $\setriI$ to losslessly produce the bit stream the worker sends to the master. 
For the lossless compression of $\ccIit$ one can employ readily available entropy coding method such as
Huffman coding or Golomb coding.

\section{Systems without error-feedback} \label{secn:withoutef}

In this section we consider momentum-SGD systems \emph{without}
error-feedback, i.e.,  when EF switch is open in \fig{block_diagram}.  The leveraging of temporal correlations to reduce communications is more straightforward for these systems, and the
insights we derive set up the algorithms we develop for error-feedback.
In \secn{quantizationdifferential} we use a linear predictor for $P$ and describe how such a predictor improves system performance. 
In \secn{numericalwithoutEF} we present numerical comparisons for systems both with and without prediction. 

\subsection{Designing the linear predictor} \label{secn:quantizationdifferential}
The goal of the predictor $P$ is to produce, in iteration $t$, an accurate estimate of $\ri[i]{t+1}$, using the $\rti[i]{t}$ vector. 
Noting that respectively from \eqn{observation_sysMod}, \eqn{errorfbri_sysMod}, \eqn{prederror_sysMod} and \eqn{errorfb_sysMod} we have 
\begin{equation}
\rti[i]{t} = \uti[i]{t} + \rhi[i]{t} = \ui[i]{t} - \ei[i]{t} + \rhi[i]{t} = \ri[i]{t} - \ei[i]{t} = \vi[i]{t} - \ei[i]{t}, \label{eqn:partialrtiandvi}
\end{equation}
and obtain the relationship between $\ri[i]{t+1}$ and $\rti[i]{t}$ as
\begin{equation}\label{eqn:noEFprediction}
\ri[i]{t+1} =\vi[i]{t+1} = \beta\vi[i]{t} + (1-\beta)\gi[i]{t+1}
= \beta\rti[i]{t} + \beta\ei[i]{t} + (1-\beta)\gi[i]{t+1},
\end{equation}
where the equalities follow from \eqn{errorfb_sysMod}, \eqn{momupdateeqn_sysMod} and \eqn{partialrtiandvi}, respectively. 

A simple, yet a good choice for $P$ is the first-order linear predictor $\fPln$ where 
\begin{equation}
\fPln(\rti[i]{t}) = \beta\rti[i]{t}. \label{eqn:predictornoEF}
\end{equation}
One may recognize that $\fPln$ in \eqn{predictornoEF} is the predictor used in the DPCM literature for first-oder Gauss-Markov sources~\cite{wiegand2011source}. 
Per the last equality in \eqn{noEFprediction}, $\ri[i]{t+1}$ is the sum of three terms. 
However, $\fPln$ in \eqn{predictornoEF} consists of only the first term of \eqn{noEFprediction}. 
This is due to two reasons. 
First, note the second term in~\eqn{noEFprediction} is the quantization error in iteration $t$. 
While there exist methods~\cite{joint_design} for designing predictors that take into account the $\ei[i]{t}$ term as well, such methods require pre-training of the predictor and the computation of the prediction is not straightforward as is $\fPln$.
As we will see later in \secn{numericalwithoutEF}, we are able to obtain large compression savings even without considering the $\ei[i]{t}$ term. 
Second, the third term in~\eqn{noEFprediction} is the stochastic gradient in iteration $t+1$. 
For large datasets and for small minibatches the correlation of $\gi[i]{t}$ across $t$ is minimal due to the randomness of the minibatch selection. 
Therefore, in practice $\gi[i]{t+1}$ will appear uncorrelated to the elements in $\rti[i]{t}$ and may not be easily estimated.

Next, let us understand how employing the predictor yields compression savings. 
To simplify the explanation let us assume that in \eqn{noEFprediction} $\ei[i]{t}$ is statistically independent of $\gi[i]{t+1}$ and $\rti[i]{t}$, and $\gi[i]{t+1}$ is uncorrelated to $\rti[i]{t}$. 
The first is a common assumption made about quantization error in mathematical analysis (e.g., in the noise shaping literature \cite{oppenheim2001discrete}). 
The second assumption is justified for large datasets and for small minibatches. 
Recall from \eqn{prederror_sysMod} that the input to the quantizer in iteration $t+1$ is $\ui[i]{t+1} = \ri[i]{t+1}-\rhi[i]{t+1}$. 
Let us consider the situations with and without prediction. 
First, if there is no prediction, i.e., if $P$ is the zero function, we have $\ui[i]{t+1} = \ri[i]{t+1}$ and the variance of $\ui[i]{t+1}$ is the sum of the variances of the three terms in~\eqn{noEFprediction}. 
Second, if we employ $\fPln$ as the predictor, we have $\ui[i]{t+1} = \beta\ei[i]{t} + (1-\beta)\gi[i]{t+1}$ and the variance of $\ui[i]{t+1}$ is the sum of the variances of only the last two terms in~\eqn{noEFprediction}. Therefore, the application of prediction reduces the variance of the quantizer input $\ui[i]{t+1}$. 

In general, we expect a quantizer input with a smaller variance to result in a smaller distortion (i.e., sum of squared components in $\ei[i]{t+1}$). 
From \eqn{sysModmain} we can write 
$\ri[i]{t+1}-\rti[i]{t+1} = \ui[i]{t+1}-\uti[i]{t+1} = \ei[i]{t+1}$. This states that the difference between $\ri[i]{t+1}$, which the worker wants to send, and $\rti[i]{t+1}$, which the master receives, is simply the quantization error $\ei[i]{t+1}$. 
Per our observation regarding the distortion, we conclude that the quantizer input with a smaller variance results in a smaller difference between $\ri[i]{t+1}$ and $\rti[i]{t+1}$. 
In summary, the temporal correlations make prediction possible, and the prediction results in a smaller difference between the vectors that the workers send and those that master receives. 
Note that $\fPln$ does not depend on the type of quantizer used in $\fQ$. 
We can plug into $\fQ$ any quantizer, including Scaled-sign, Top-$K$ and Top-$K$-Q, 
and employ $\fPln$ as the predictor.

\subsection{Numerical comparisons} \label{secn:numericalwithoutEF}

\begin{figure*}
\centering\includegraphics[width=0.9\textwidth]{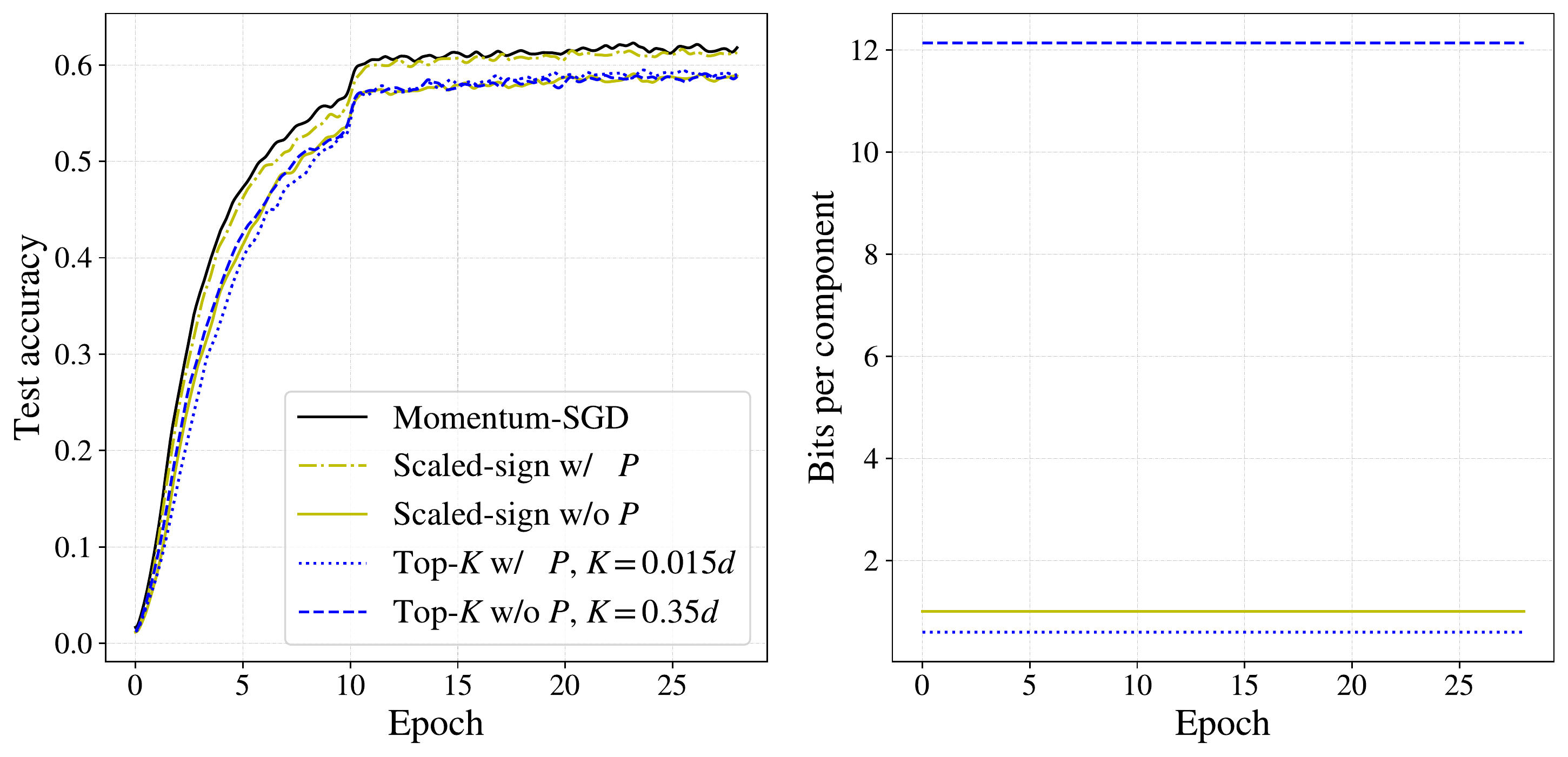}
\caption{
Comparing the performance of Scaled-sign and Top-$K$ with (w/) and without (w/o) the predictor when error-feedback is not used. 
We employ $\fPln$ discussed in \secn{withoutef} as the predictor $P$. All algorithms use momentum with $\beta=0.99$. 
The left and right sub-figures plot top-5 test accuracy and number of bits per gradient component. 
In the right sub-figure the two plots for Scaled-sign experiments overlap and so the two plots are not distinguishable.
} 
\label{fig:numericalimagenetcomparecsgd}
\end{figure*}

\begin{figure*}
\centering\includegraphics[width=0.9\textwidth]{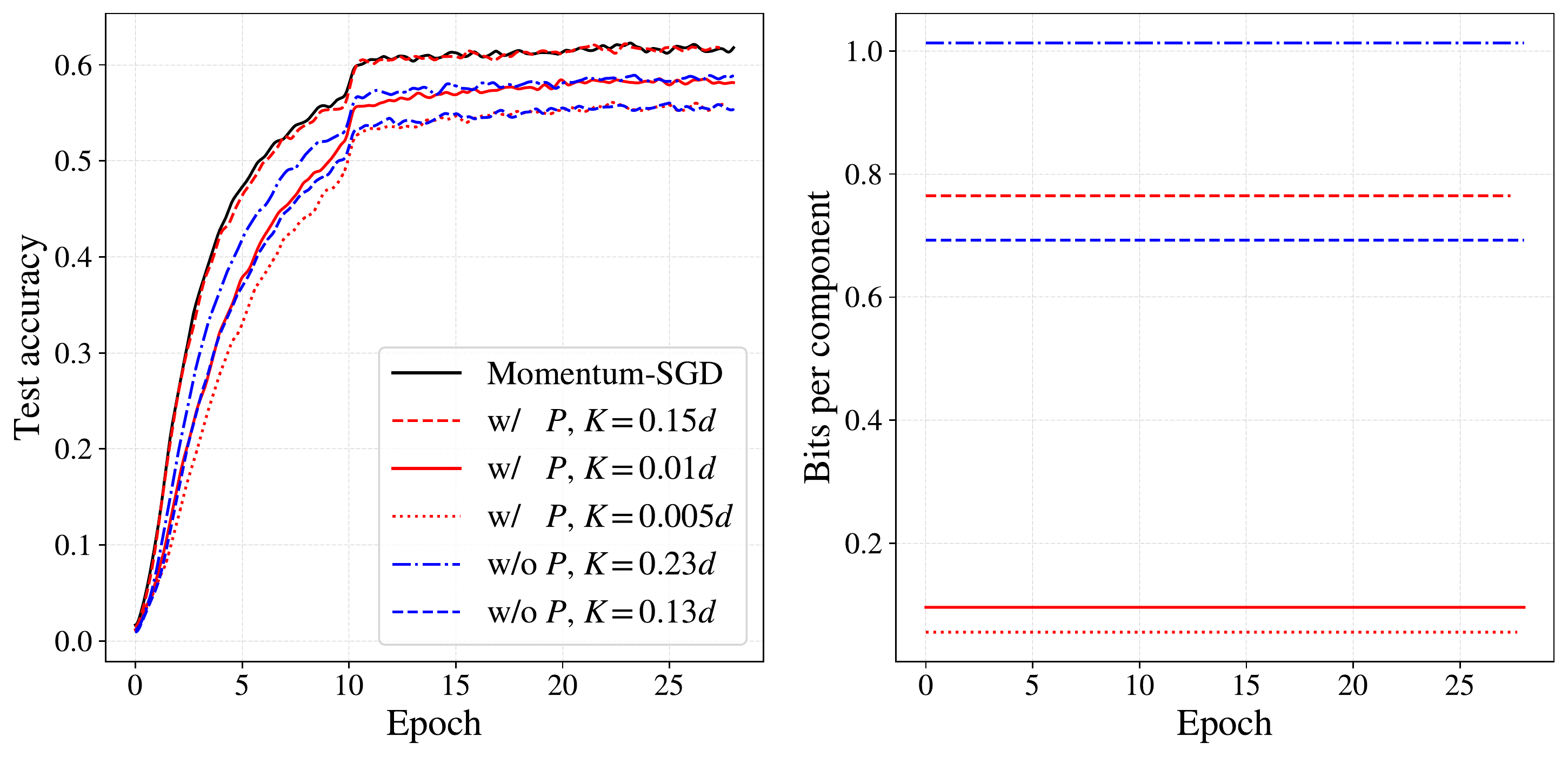}
\caption{
This figure is similar to \fig{numericalimagenetcomparecsgd}. 
In all plots except momentum-SGD the quantizer employed is Top-$K$-Q, the Top-$K$ variant with quantization considered in \cite{dryden2016communication}. 
} 
\label{fig:numericalcifcomparecsgd}
\end{figure*}

In Figs.~\ref{fig:numericalimagenetcomparecsgd} and~\ref{fig:numericalcifcomparecsgd} we compare the performance of different quantizers with and without the $\fPln$ predictor. 
Results are obtained by training a convolutional neural network (WRN-28-2) on a master-worker setup using 4 workers. 
The details of the experiments are noted in \secn{numericaliclrtemporal}. 
In each algorithm the encoder $\mathcal{E}$ losslessly compresses the $\uti[i]{t}$ vector. 
The number of bits per vector component is calculated as follows. 
Momentum-SGD uses $32$ bits per component. 
For Scaled-sign we have a binary vector of dimension $\dimw$ that encodes the signs of the compressed vector. 
In our experiments we observed that this binary vector consists of roughly the same amounts of ones and zeros, therefore, it is not further compressible and takes 1-bit per component. 

Recall that with the Top-$K$ quantizer only $K$ components in $\uti[i]{t}$ are non-zero. 
We use a binary vector of dimension $\dimw$ to encode the locations of non-zero components. 
For large $\dimw$ this vector can be losslessly compressed with close to $\dimw H_b(K/d)$ bits, where $H_b$ is the binary entropy function. 
Since $\dimw$ is well over a million for models of practical interest,
there exist readily available lossless compression algorithms such as
Huffman coding or Golomb coding that can compress losslessly
the binary vector to rates very close to its Shannon entropy.  
For example, in \cite{strom2015scalable}
and \cite{sattler2019sparse} authors use Golomb coding to encode
non-zero locations of the Top-$K$ compression.  
Together with the compressed binary vector and the number of bits required to describe the $K$ non-zero values, Top-$K$ takes $H_b(K/\dimw) + 32K/\dimw$ bits per component.

In \fig{numericalimagenetcomparecsgd} we compare 
Scaled-sign and Top-$K$ quantizers with and without prediction. 
The experiments with the predictor easily outperforms the ones without the predictor. 
For example, Scaled-sign without the predictor only reaches an end accuracy of around $58.5\%$, whereas 
Scaled-sign with the predictor closely matches the momentum-SGD (with no compression so, using floating-point, requires $32$ bits per component) accuracy (left sub-figure) while
requiring the same communication payload of one bit per component (right sub-figure).
To match the same accuracy of $58.5\%$, Top-$K$ without prediction requires an even higher payload,
about $12$ bits per component, c.f., \tbl{topklsummary} as well. 
Indeed to recover the same test accuracy, Top-$K$ with prediction requires only $0.6$ bits per component.

In \fig{numericalcifcomparecsgd} we also compare the impact of prediction with the Top-$K$-Q quantizer, 
which is a variant of Top-$K$ introduced in~\cite{dryden2016communication}. 
Top-$K$-Q is similar to Top-$K$, except that non-zero values are
quantized.  All positive non-zero values and all negative non-zero
values belong to two separate reconstruction points. 
The association of each component in $\uti[i]{t}$ to a reconstruction point can be represented by a ternary vector. 
We calculate the size of the communication payload by computing the entropy of the ternary vector. 
We present the comparisons with Top-$K$-Q in \fig{numericalcifcomparecsgd}. 
Observe that the test accuracy of Top-$K$-Q with prediction and $K=0.005\dimw$, and Top-$K$-Q without prediction and $K=0.13\dimw$ are nearly the same, although the latter uses around $0.7$ bits 
whereas the former uses only around $0.05$ bits per component. 
While compression with prediction should offer even more savings with a $\beta$ larger than $0.99$, we observed that the test accuracy degrades when $\beta$ is made larger (gradient is smoothed too much).

We again remind the reader that these results are tabulated in the
second section of \tbl{topklsummary}. In summary, the results in
Figs.~\ref{fig:numericalimagenetcomparecsgd}
and~\ref{fig:numericalcifcomparecsgd} demonstrate a significant
opportunity for exploiting temporal correlation to effect reduction in
communication.

\section{Systems with error-feedback} \label{secn:algorithmef}

We now extend our ideas of leveraging temporal correlation in momentum
vectors to systems that use error-feedback. The use of error-feedback
by itself improves compression rates while keeping performance nearly
the same as that realized without error-feedback. Another advantage to
error-feedback is that $\delta$-compressors such as Top-$K$ and
Scaled-sign are guaranteed to converge when error-feedback is
employed \cite{stich2018sparsified, karimireddy2019error,
zheng2019communication}.  However, exploitation of correlated momentum
vectors with error-feedback requires a more careful treatment of the
problem than was needed in \secn{withoutef}. In \secn{compprobef} we
discuss why the linear prediction scheme discussed in \secn{withoutef} is not very useful when error-feedback
is employed. In \secn{illustrative} we present an illustrative 
example that motivates our proposed Est-$K$ predictor. 
Building on the insights we derive in \secn{illustrative}, 
in \secn{illustrativeestk} we propose our Est-$K$ (``estimate-top-$K$'') algorithm which exploits temporal correlation in systems that use error-feedback.  
The Est-$K$ predictor works in conjunction with the Top-$K$ quantizer. 
We choose to build on Top-$K$ as it already realizes large
compression ratios in the context of error-feedback and closely
matches the performance of momentum-SGD (without compression).
In \secn{EFB_numresults} we present experimental results comparing
momentum-SGD to Top-$K$ with and without prediction.

\subsection{Linear prediction with error-feedback} \label{secn:compprobef}
In this section we discuss why the linear predictor $\fPln$ in~\eqn{predictornoEF} is not a good predictor for systems that employ error feedback. 
First, recalling that $\fPln(\rti[i]{t}) = \beta\rti[i]{t}$, let us interpret its formulation in the following manner. 
From \eqn{noEFprediction} we have 
\begin{equation}
\ri[i]{t+1} =
\beta\vi[i]{t} + (1-\beta)\gi[i]{t+1} = 
\overbrace{\underbrace{\beta\rti[i]{t}}_{\mathclap{\fPln(\rti[i]{t})}} 
+ \beta\ei[i]{t}}^{\mathclap{\beta\vi[i]{t}}} + (1-\beta)\gi[i]{t+1}. \label{eqn:equalitycopynoEF}
\end{equation}
We observe that the momentum term $\vi[i]{t}$ is the sum of $\rti[i]{t}$ and $\ei[i]{t}$. 
Ignoring the error term $\ei[i]{t}$ as quantization noise, we can consider $\rti[i]{t}$ to be an estimate of $\vi[i]{t}$. 
The predictor uses this estimate of $\vi[i]{t}$ to predict $\ri[i]{t+1}$. 
Observing in the first equality of \eqn{equalitycopynoEF} that the contribution of $\vi[i]{t}$ to $\ri[i]{t+1}$ is $\beta\vi[i]{t}$, the predictor produces the prediction by weighting the estimate of $\vi[i]{t}$ by $\beta$, yielding the formulation of $\fPln$ in \eqn{predictornoEF}. 
Therefore, the role of the predictor can be interpreted as estimating the momentum term $\vi[i]{t}$ and weighting it appropriately to predict $\ri[i]{t+1}$. 

Now, let us turn to systems that employ error-feedback, i.e., systems with the EF switch in \fig{block_diagram} closed. 
To simplify the explanation let us assume that the learning rate $\stepsize_{t}$ is a constant. 
In systems with error feedback we have 
\begin{equation}
\rti[i]{t} = \uti[i]{t} + \rhi[i]{t} = \ui[i]{t} - \ei[i]{t} + \rhi[i]{t} = \ri[i]{t} - \ei[i]{t} = \vi[i]{t} + \ei[i]{t-1} - \ei[i]{t}, \label{eqn:errorline}
\end{equation}
where we obtain each equality by substituting from~\eqn{sysModmain} appropriately. 
This gives 
\begin{equation}
\beta\vi[i]{t} = \underbrace{\beta\rti[i]{t}}_{\mathclap{\fPln(\rti[i]{t})}} + \beta\ei[i]{t} -\beta\ei[i]{t-1}. \label{eqn:partialrtiandvireason}
\end{equation}
The last term in \eqn{partialrtiandvireason} is due to error feedback. 
Compared to \eqn{equalitycopynoEF}, we observe that in \eqn{partialrtiandvireason} if we were to approximate $\beta\vi[i]{t}$ by $\fPln(\rti[i]{t})$, the resulting approximation error would not only be contributed by the quantization noise $\ei[i]{t}$ in iteration $t$, but also by $\ei[i]{t-1}$ that corresponds to iteration $t-1$. 
It can be demonstrated that $\ei[i]{t}$ grows unbounded with $t$ if we were to use $\fPln$ as the predictor. 

\begin{figure}
\centering\includegraphics[width=0.45\textwidth]{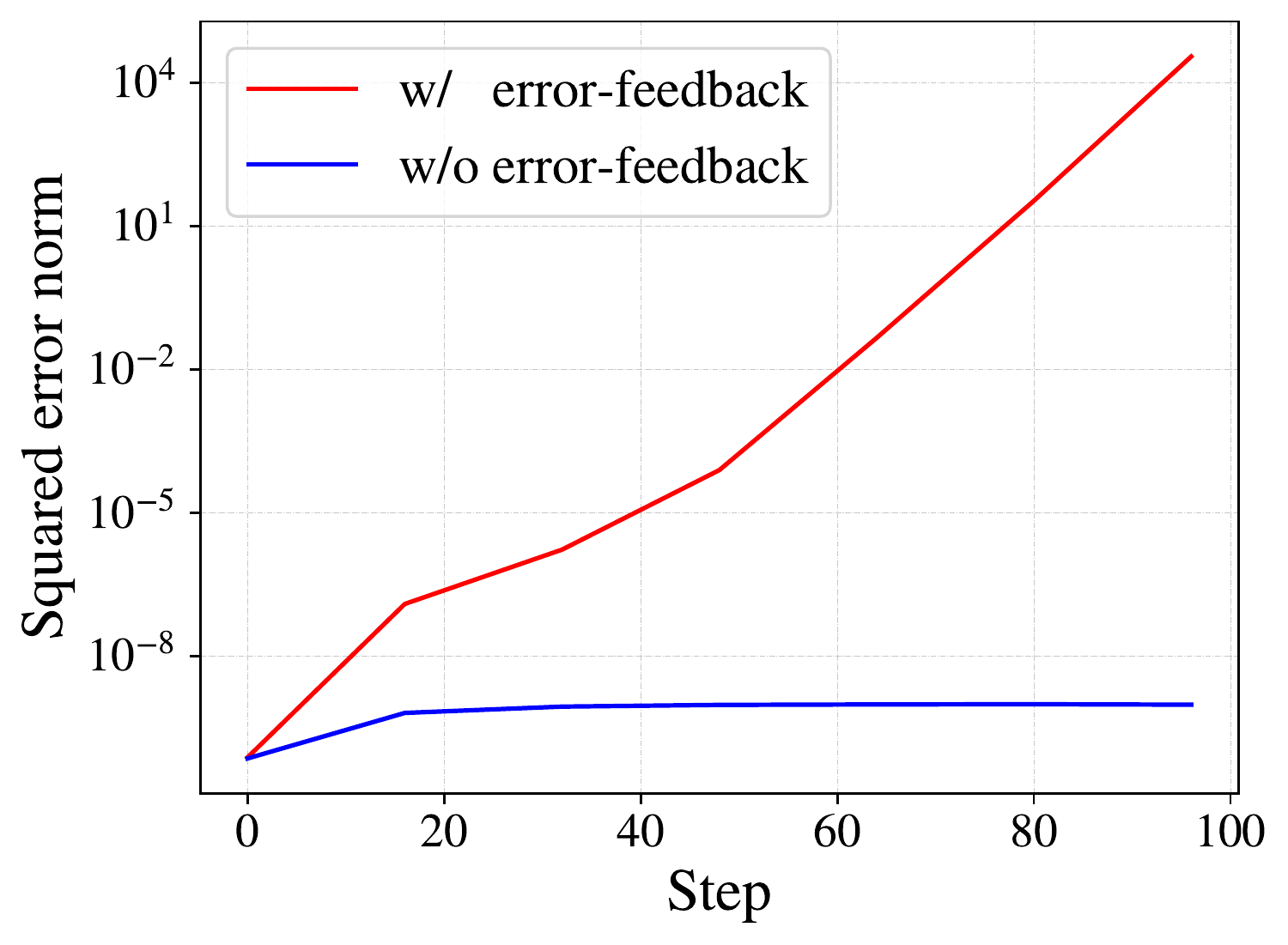}
\caption{The evolution of $\Lnrm{\ei[i]{t}}^2$ in the first 100 iterations with Top-$K$-Q quantizer and $\fPln$ predictor.}
\label{fig:imagenet32_linear_problem}
\end{figure}

\fig{imagenet32_linear_problem} shows the evolution of the squared quantization error norm $\Lnrm{\ei[i]{t}}^2$ when the predictor is $\fPln$ and the quantizer is Top-$K$-Q. 
The details of the experimental setup are noted in \secn{numericaliclrtemporal}. 
The only difference between the two plots in \fig{imagenet32_linear_problem} is that 
they correspond to systems with and without error-feedback. 
We observe that in the system that employs error-feedback the error norm grows unbounded. 
This is undesirable since from \eqn{errorline} we have $\ei[i]{t} = \ri[i]{t}-\rti[i]{t}$, 
which states that $\ei[i]{t}$ is the difference between the vector worker wants to send and the vector the master receives. 
With this observation we conclude that when error-feedback is employed 
$\fPln(\rti[i]{t})$ is not a good prediction of $\ri[i]{t+1}$.

\subsection{An illustrative example that motivates Est-$K$} \label{secn:illustrative}
To present the insight behind our proposed Est-$K$ predictor we first consider
the error-feedback system with a master and a single worker that
mimics momentum-SGD with Top-$K$ compression.  With reference
to \fig{block_diagram} we set $Q$ to $\fQT$ and $P$ to the zero function (so
$\rhi{t}=\allzeros$ and $\ui{t}=\ri{t}$). 
Recall that as we describe in \secn{exampletopk} the $\fQT$ quantizer sparsifies the input. 
To summarize,
\begin{align}
\vi{t} &= \beta\vi{t-1} + (1-\beta)\gi{t} \label{eqn:momupdateeqn} \\
\ui{t} &= \ri{t} = \vi{t} + \ei{t-1} \label{eqn:errorfb} \\
\uti{t} &= \rti{t} = \fQT(\ui{t}) \label{eqn:errorfbQ} \\
\ei{t} &= \ui{t} - \uti{t}. \label{eqn:errorfbri}
\end{align}
Although in error-feedback we typically scale $\ei{t}$ to adjust for
the varying steps size $\eta_t$, here we assume that step size is a
constant and ignore the scaling.  This helps simplify the example.
The operator $\fQT$ produces the sparse vector $\uti{t}$ as follows.  
Let $\ccIitp{t}$ be the set of $K$ indices in
$\ui{t}$ largest in magnitude.  We have $\uti{t}[k]=\ui{t}[k]$ if
$k\in\ccIitp{t}$, and $\uti{t}[k]=0$ if $k\notin\ccIitp{t}$.

\begin{figure*}
\centering\includegraphics[width=0.95\textwidth]{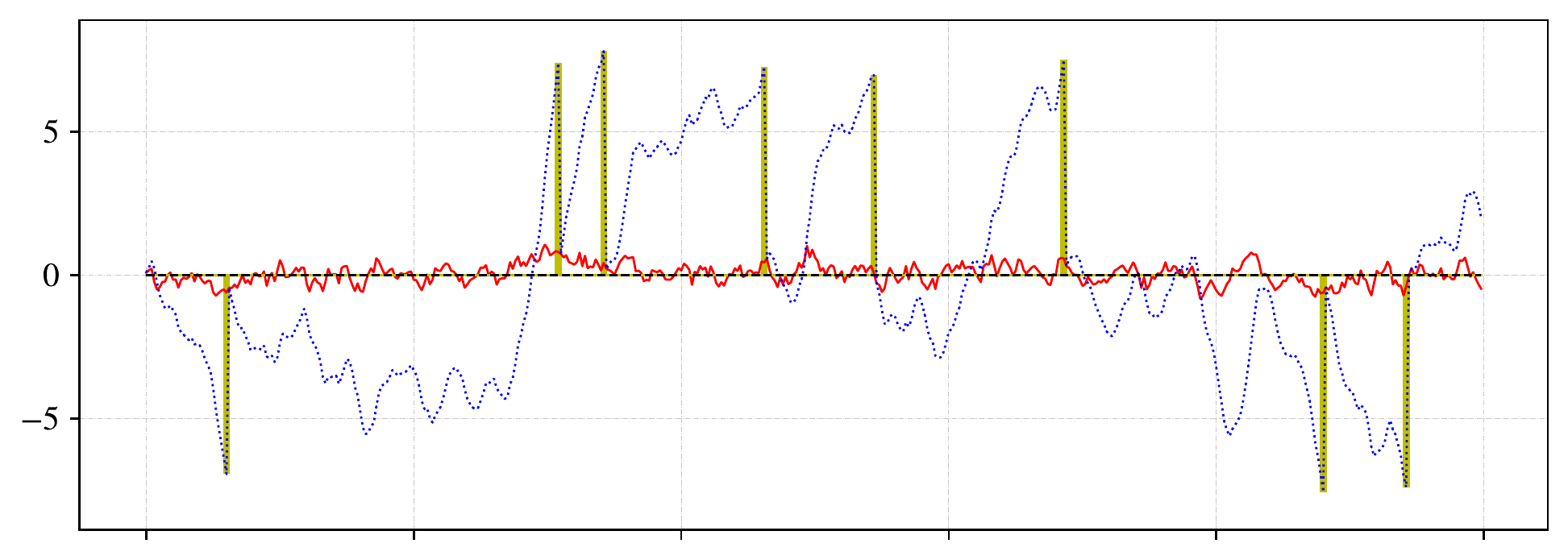}
\centering\includegraphics[width=0.95\textwidth]{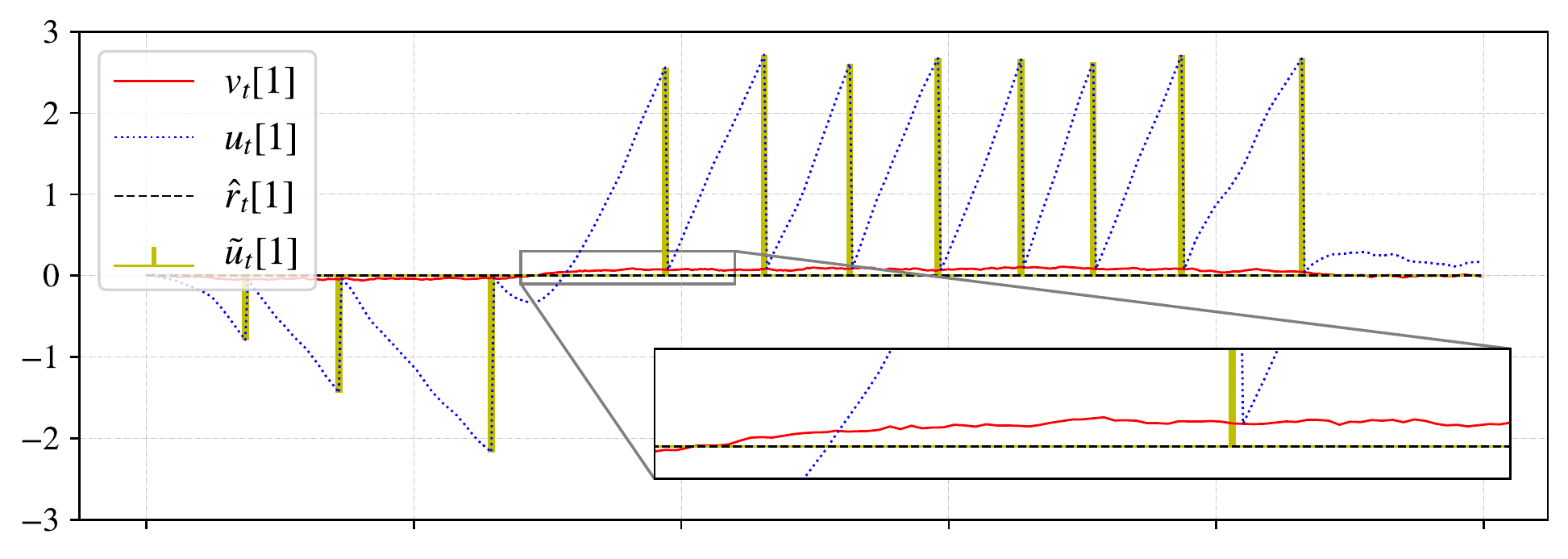}
\centering\includegraphics[width=0.95\textwidth]{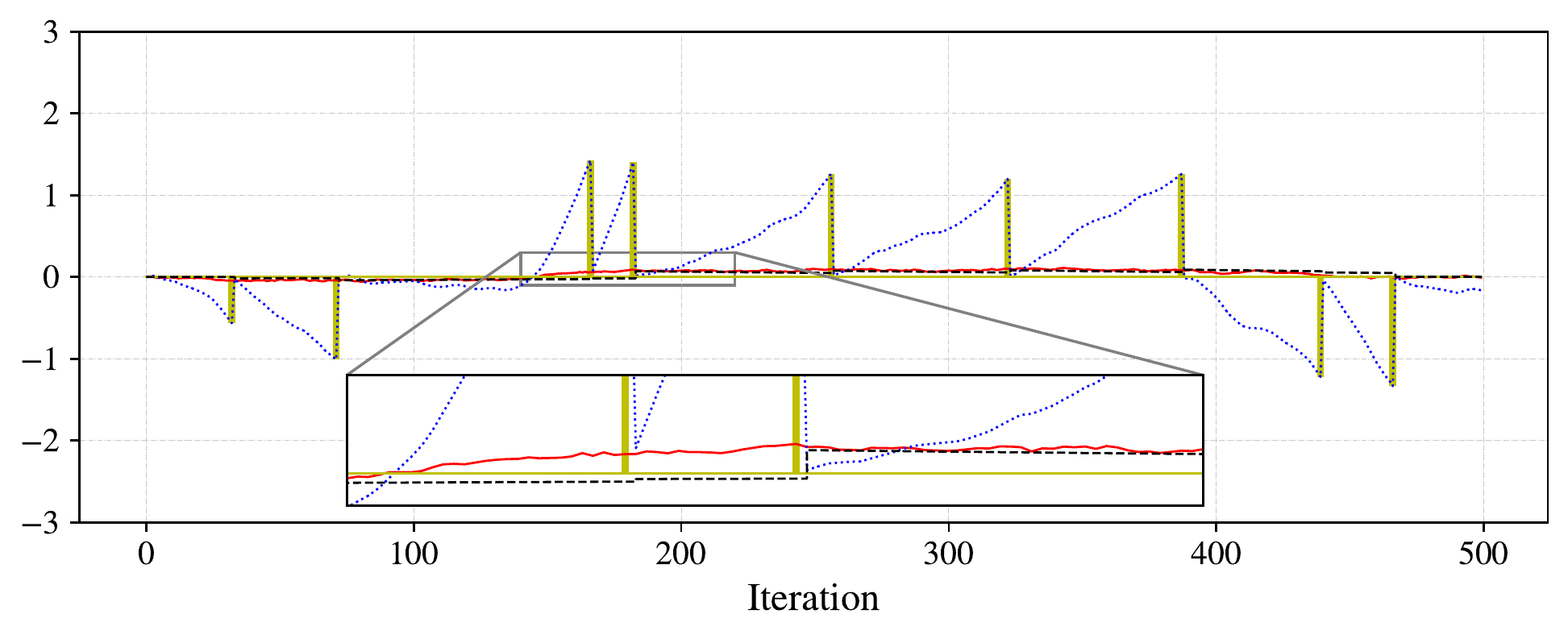}
\caption{
The iterates in the synthetic experiment. 
\secn{illustrative} discusses top (a) and middle (b) sub-figures, which are generated with $\beta=0.8$ and $\beta=0.995$ respectively using $\fQT$ without prediction. 
{The bottom sub-figure (c) is discussed in \secn{illustrativeestk}. It
is generated with $\beta=0.995$ using $\fQT$ with the predictor Est-$K$.}  In all three cases
$\gi{t}$ is sampled using the same random seed.  Therefore,
$\vi{t}[1]$ in (b) and (c) are identical.
}
\label{fig:time_series}
\end{figure*}

Our goal is to understand how a randomly selected component of
$\vi{t}$, $\ui{t}(=\ri{t})$ and $\uti{t}$ evolves.  To this
end we present the following synthetic experiment.  We mimic the
gradient $\gi{t}$ by sampling its components independently from the
standard normal distribution.  In models of the practical interest,
such as ResNet-20 and ResNet-50, the distribution of the components in
the gradient is uni-modal and symmetric \cite{glorot2010understanding,
bernstein2018signsgd, shi2019understanding}.  Therefore, this
synthetically generated $\gi{t}$ is consistent with empirical
observations.  We set $\dimw=1000$ and $K=0.01\dimw$.  Since the
sampling of the gradient is independent, the first component is as
good as any other randomly selected component.  

In Figs.~\ref{fig:time_series}(a) and (b), respectively the top and
middle subfigures, we plot the first component of each vector
($\vi{t}[1]$, $\ui{t}[1]$ etc.) versus iteration index $t$.  In
\fig{time_series}(a) the bandwidth parameter $\beta = 0.8$
while in \fig{time_series}(b) $\beta = 0.995$.  As one may
expect, a larger $\beta$ yields a smoother $\vi{t}[1]$.  A peak
in $\uti{t}[1]$ means that in this particular iteration $\ui{t}[1]$
is one of the $K$ values of largest magnitude, i.e.,
$1\in\ccIitp{t}$.  An important observation is that for large $\beta$
(\fig{time_series}(b)) the spacing between peaks of $\uti{t}[1]$ 
is much more consistent (occurs in regular intervals) than that for a small $\beta$ (\fig{time_series}(a)). 
Consider an interval in \fig{time_series}(b)
where $\vi{t}[1]$ is comparatively large and remains stable, such as
from iteration $200$ to $400$.  The magnitude of $\ui{t}[1]$
grows between every two adjacent peaks of $\uti{t}[1]$.  When
$\ui{t}[1]$ gets large enough it gets sent to master, resulting in a peak
in $\uti{t}[1]$.

Our goal is to design an algorithm for $P$ that predicts $\ri{t+1}= \vi{t+1} + \ei{t}$ using $\rti{t}$. 
Note that the predictor can use $\rti{t}$ to recover the sparse vector $\uti{t}$. 
Our idea is that we can use the non-zero instances of $\uti{t}$ 
to estimate and extrapolate $\vi{t}$, thereby estimating one of the two terms that make up $\ri{t}$. 
In comparison to estimating $\vi{t}$ with a small $\beta$ (\fig{time_series}(a)), estimating and extrapolating $\vi{t}$ with a large $\beta$ (\fig{time_series}(b)) will be more accurate since in the latter $\vi{t}$ changes only slowly across the iterations.
An accurate estimate of $\vi{t}$ leads to a good prediction of $\ri{t}$. 
We propose in \algo{encdecestk} a simple algorithm for the predictor $P$ 
when the quantizer in \fig{block_diagram} is $\fQT$. 
We refer to this predictor as Est-$K$ to reflect the estimating aspect of \algo{encdecestk} and to indicate its association with $\fQT$.

\subsection{Designing the Est-$K$ predictor} \label{secn:illustrativeestk}
To use the Est-$K$ predictor we set the functioning of $P$ in \fig{block_diagram} 
to \algo{encdecestk}.  For the rest of the system we make the same
choices for $Q$, $\mathcal{E}$ and $\mathcal{D}$ as we do for Top-$K$.
For each worker the master operates a separate decoding-and-prediction
chain composed of a $\calD$, a $P$, and a delay block. In \algo{encdecestk} the
index $i$ refers to the worker index.  As is illustrated
in \fig{block_diagram}, worker-$i$ also locally operates a predictor block.
The master's $i$th decode-and-predict chain thereby keeps in sync
with worker-$i$'s prediction. 

The input and output of \algo{encdecestk} are $\rti[i]{t}$ and $\rhi[i]{t+1}$ as defined in \eqn{errorfbQPredict_sysMod}. 
Note that $\ccIit$ is the set of $K$ indices in $\ui[i]{t}$ largest in magnitude. 
This set can be recovered using $\uti[i]{t}$ as done in \algo{encdecestk}. 
In vector $\taui[i]{t}$, $\taui[i]{t}[k]$ is the
number of iterations since the last time master received a non-zero
update for the $k$th component.  In other words, the number of
iterations since the last time index $t$ at which the master received
a set $\ccIit$ such that $k\in\ccIit$. 
The vector $\vhi[i]{t}$ in \algo{encdecestk} keeps track of the last estimate of the momentum vector. 

Consider an index $k\in[\dimw]$. 
In summary, \algo{encdecestk} estimates the momentum $\vi[i]{t}[k]$ between every two peaks of $\uti[i]{t}[k]$ (see \fig{time_series}(b)) by computing the average rate of change of $\vi[i]{t}[k]$ within that interval. 
If $\vi[i]{t}[k]$ changes only slowly, such as the case from iteration $200$ to $400$ in \fig{time_series}(b), then the average rate of change is a good approximation of $\vi[i]{t}[k]$. This can be extrapolated as the approximation of $\vi[i]{t}[k]$ in next iterations. 

\begin{algorithm}
\caption{Operation of the Est-$K$ predictor} \label{algo:encdecestk}
\SetAlgoVlined \LinesNumbered \SetKwInOut{Initialize}{initialize}
\SetKwInOut{Input}{input} \SetKwInOut{Output}{output}
\Initialize{$\vhi[i]{0}=\allzeros$, $\rhi[i]{0}=\allzeros$ and $\taui[i]{0}=\allzeros$ \;}
\Input{$\rti[i]{t}$ \;}
\Output{$\rhi[i]{t+1}$ \;}
recover $\uti[i]{t} = \rti[i]{t} - \rhi[i]{t}$ and 
obtain $\ccIit$, the set of non-zero indices in $\uti[i]{t}$ \;
\For{$k\in[\dimw]$}{
$\vhi[i]{t+1}[k] = \begin{cases}
\frac{(\beta+\dots+\beta^{\taui[i]{t}[k]+1})\vhi[i]{t}[k] + \uti[i]{t}[k]}{\taui[i]{t}[k]+1} & \text{if $k\in\ccIit$}\\
\vhi[i]{t}[k] & \text{if $k\not\in\ccIit$}
\end{cases}$ \;
~\\
compute $\taui[i]{t+1}[k] = \begin{cases}
0 & \text{if $k\in\ccIit$}\\
\taui[i]{t}[k]+1 & \text{if $k\not\in\ccIit$}
\end{cases}$ \;
~\\
compute $\rhi[i]{t+1}[k] = \beta^{\taui[i]{t+1}[k]+1} \vhi[i]{t+1}[k]$
}
\end{algorithm}

\begin{table*}
\renewcommand{\arraystretch}{1.4}
\caption{An example of the evolution of iterates with Est-$K$.}
\label{tbl:iteratesestk}
\centering
\begin{tabular}{|c|l|c|c|c|c|c|c|r|}
\hline
{$t$}        														&
\multicolumn{1}{c|}{$\ri{t} = \vi{t} + \ei{t-1}$} 					&
$\rhi{t}$ 															&
$\ui{t}=\ri{t}-\rhi{t}$ 											&
$\uti{t}$ 															&
$\rti{t}=\uti{t}+\rhi{t}$ 											&
\multicolumn{1}{c|}{$\ei{t} = \ri{t} - \rti{t}$} 					&
$\taui{t}$ 															&
\multicolumn{1}{c|}{$\vhi{t}$}									\\ \hline\hline
$0$ & $\vi{0}$ 								& $0$ 					& $\ri{0}$ 					& $0$ 					& $0$ 					& $\ui{0}$ 						& $0$ & ${0}$ \\
$1$ & $\vi{1} + \vi{0}$ 					& $0$ 					& $\ri{1}$ 					& $0$ 					& $0$ 					& $\ui{1}$			 			& $1$ & ${0}$ \\
$2$ & $\vi{2} + \vi{1} + \vi{0}$ 			& $0$ 					& $\ri{2}$ 					& $0$ 					& $0$ 					& $\ui{2}$ 						& $2$ & ${0}$ \\
$3$ & $\vi{3} + \vi{2} + \vi{1} + \vi{0}$ 	& $0$ 					& $\ri{3}$ 					& {$\ui{3}$}		& $\ri{3}$ 				& $0$ 							& $3$ & $\frac{0+\uti{3}}{4} = \frac{\vi{3} + \vi{2} + \vi{1} + \vi{0}}{4}$ \\
$4$ & $\vi{4}$ 								& $\beta\vhi{3}$ 		& $\ri{4}-\beta\vhi{3}$		& $0$ 					& $\beta\vhi{3}$ 		& $\ui{4}$ 						& $0$ & $\vhi{3}$ \\
$5$ & $\vi{5} + \vi{4}-\beta\vhi{3}$		& $\beta^2\vhi{3}$ 		& $\ri{5}-\beta^2\vhi{3}$	& $0$ 					& $\beta^2\vhi{3}$ 		& $\ui{5}$ 						& $1$ & $\vhi{3}$ \\
$6$ & $\vi{6} + \vi{5} + \vi{4}-\beta\vhi{3}-\beta^2\vhi{3}$		
& $\beta^3\vhi{3}$ 		& $\ri{6}-\beta^3\vhi{3}$	& {$\ui{6}$} 		& $\ri{6}$ 				& $0$ 							& $2$ & $\frac{(\beta+\beta^2+\beta^3)\vhi{3} + \uti{6}}{3} = \frac{\vi{6} + \vi{5} + \vi{4}}{3}$ \\
$7$ & $\vi{7}$ 								& $\beta\vhi{6}$ 		& $\ri{7}-\beta\vhi{6}$ 	& $0$ 					& $\beta\vhi{6}$ 		& $\ui{7}$ 						& $0$ & $\vhi{6}$ \\
$\vdots$&$\vdots$&$\vdots$&$\vdots$&$\vdots$&$\vdots$&$\vdots$&$\vdots$& \\
\hline
\end{tabular}
\end{table*}

\algo{encdecestk} can best be understood by considering one component (e.g., the $k$th component) in each iterate $\ri[i]{t}, \rhi[i]{t}, \ui[i]{t}, \uti[i]{t}, \rti[i]{t}, \ei[i]{t}, \taui[i]{t}$ and $\vhi[i]{t}$, for the first few iterations $t=\{0,1,2,\dots\}$. In \tbl{iteratesestk} we present an example by tabulating these iterates. To avoid clutter, we do not indicate the worker index $i$ and the component index $k$ of the iterates. We assume that the learning rate $\stepsize_{t}$ is a constant. Recall that the quantizer employed with Est-$K$ is $\fQT$. The iterations where $\uti{t}$ is non-zero in \tbl{iteratesestk} are the iterations for which $\ui{t}[k]$ is one of the $k$ values in $\ui{t}$ with largest magnitudes, i.e., when $k\in\ccIit$ and $\fQT$ sets $\uti{t}[k]=\ui{t}[k]$. 
For $t=3$ the algorithm computes $\vhi{3} = \frac{\vi{3} + \vi{2} + \vi{1} + \vi{0}}{4}$. If the momentum vector evolves only slowly from $t=0$ to $t=3$, then $\vhi{3}$ is a good approximation of $\vi{3}$, which can be used as the estimate of $\vi{3}$ in the subsequent iterations. For $t=4$ we have $\ri{4} = \vi{4} = \beta\vi{3} + (1-\beta)\gi{4}$. In this case $\rhi{4}=\beta\vhi{3}$ can be used as the approximation of the $\beta\vi{3}$ term in $\ri{4}$. Similarly, for $t=5$ we can expand and write $\ri{5}$ as
$ \ri{5}
= \beta^2\vi{3} + \beta(\vi{3}-\vhi{3}) + (1-\beta^2)\gi{4} + (1-\beta)\gi{5}
$. 
In this case $\rhi{5}=\beta^2\vhi{3}$ approximates the $\beta^2\vi{3}$ term in $\ri{5}$.

Let us revisit the illustrative example in \secn{illustrative} and
apply Est-$K$.  In this case we can disregard $i$ as we only have one
worker.  Since $P$ is no longer the zero function, now $\ri{t}$ and
$\ui{t}$ are not necessarily equal.  We use the same $\beta$ and $K$
values, and the same sequence of $\gi{t}$ as in \fig{time_series}(b).
The resulting iterates are presented in \fig{time_series}(c).  Note
that the sample paths for $\vi{t}[1]$ in \fig{time_series}(b)
and \fig{time_series}(c) are identical.  We observe that
$\rhi{t}[1]$ follows $\vi{t}[1]$ closely with Est-$K$ (see the magnified duration). 
The maximum magnitude of $\ui{t}[1]$ with Est-$K$ is around half that of Top-$K$. 
This demonstrates that the predictor in Est-$K$ lowers the variance of $\ui{t}[1]$, the input to the quantizer.

Finally, we quickly summarize the extension to a multi-worker system,
i.e., when $\numDists > 1$.  \algo{momentumSGDEF} formally extends the Est-$K$
scheme to such settings.  Note that the definitions of iterates
in \algo{momentumSGDEF} are consistent with those
in \fig{block_diagram}.  In line \ref{lst:line:topksend} the payload sent to the master consists of the bit stream that encodes $\ccIit$ and $\setriI$.

\begin{algorithm} 
\caption{Extension of the Est-$K$ method to multi-worker setting} \label{algo:momentumSGDEF}
\SetAlgoVlined \LinesNumbered \SetKwInOut{Initialize}{initialize}
\SetKwInOut{Input}{input} \SetKwInOut{Output}{Output}
\Input{parameter vector $\wi{0}$ ; 
step sizes $\{\stepsize_0, \dots, \stepsize_{T-1}\}$; 
$0\leq\beta<1$ \; 
}

\Initialize{
\textbf{on master} set $\rhi[i]{0}=\allzeros$ and initialize \algo{encdecestk} for all $i\in[\numDists]$ ; \\
\textbf{on worker-$i$} initialize \algo{encdecestk}, set $\rhi[i]{0}=\allzeros$, \\
$\vi[i]{-1}=\allzeros$, $\ei[i]{-1}=\allzeros$ and $\stepsize_{-1}=0$ \;
}
\BlankLine
\For{$t\in\{0,\dots,T-1\}$}{ 
\SetKwBlock{workercode}{\textbf{on worker-$i$}}{}
\workercode{
compute $\gi[i]{t}$, the stochastic gradient at $\wi{t}$ \;
set $\vi[i]{t} = \beta\vi[i]{t-1} + (1-\beta)\gi[i]{t}$ \;
set $\ri[i]{t} = \vi[i]{t} + \frac{\stepsize_{t-1}}{\stepsize_{t}}\ei[i]{t-1}$ \;
compute $\ui[i]{t} = \ri[i]{t}-\rhi[i]{t}$ \;
apply $\fQT$ to $\ui[i]{t}$ and obtain $\uti[i]{t}$ \;
encode $\uti[i]{t}$ and send the result to the master \label{lst:line:topksend} \;
set $\rti[i]{t} = \uti[i]{t} + \rhi[i]{t}$ \;
compute $\rhi[i]{t+1}$ as per \algo{encdecestk} \;
set $\ei[i]{t} = \ri[i]{t}-\rti[i]{t}$ \;
receive $\avgopn\rti[i]{t}$ from master \;
update $\wi{t+1} = \wi{t} - \stepsize_{t}\avgopn\rti[i]{t}$ \;
}

\BlankLine\BlankLine
\SetKwBlock{mastercode}{\textbf{on master}}{}
\mastercode{
\For{$i\in[\numDists]$}{
receive the encoded $\uti[i]{t}$ from worker-$i$ and decode to obtain $\uti[i]{t}$ \;
set $\rti[i]{t} = \uti[i]{t} + \rhi[i]{t}$ \;
compute $\rhi[i]{t+1}$ as per \algo{encdecestk} \label{lst:line:topkdiffmaster} \;
}
broadcast $\avgopn\rti[i]{t}$ to all workers \;
}
} 
\end{algorithm}

\subsection{Numerical comparisons} \label{secn:EFB_numresults}

\begin{figure*}
\centering\includegraphics[width=0.9\textwidth]{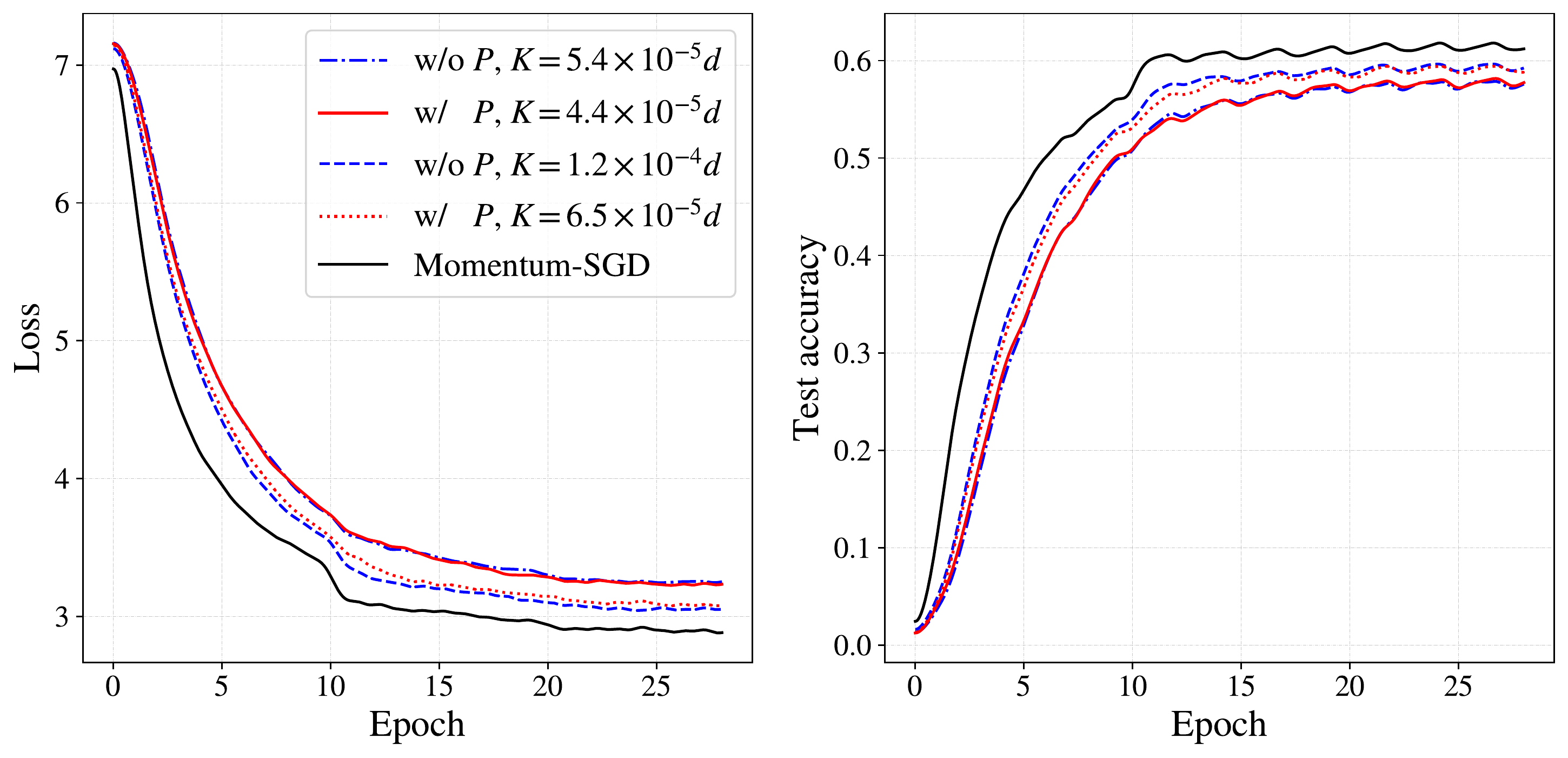}
\caption{Comparing the performance of Est-$K$ with Top-$K$. 
All algorithms employ momentum with $\beta=0.99$ parameter. Est-$K$ and Top-$K$ employ error-feedback. 
} 
\label{fig:imagenet32_EFSGD1w}
\end{figure*}

\fig{imagenet32_EFSGD1w} presents a performance comparison between momentum-SGD (without compression), Top-$K$ without a predictor, and Top-$K$ with the Est-$K$ predictor. 
Experiments with Top-$K$ compression employ error-feedback. 
Since $K$ is common to experiments with compression $K$ serves as the parameter that parameterizes compression savings. 
Results are obtained by training a wide
residual network classifier on the ImageNet-$32$ dataset using $4$
workers.  Details of our experimental setup are discussed
in \secn{numericaliclrtemporal}. 
Recall that in \fig{numericalcifcomparecsgd} (experiments without error-feedback) we presented the performance of Top-$K$-Q. 
Since the Top-$K$ compression ratio is high, i.e., the
ratio $K/\dimw$ is small, the compression performance of Top-$K$-Q is very close to Top-$K$. 
(When $K$ is small quantizing the non-zeros values in Top-$K$ has only
a small impact on the overall rate.)  Therefore, we do not include
Top-$K$-Q in our comparison with error-feedback. 

As observed in \fig{imagenet32_EFSGD1w} we tune $K$ in each of experiment to reach two test accuracy levels. 
Top-$K$ without prediction reaches the lower test accuracy level with $K=5.4\times10^{-5}\dimw$, 
whereas Top-$K$ with prediction requires only around $K=4.4\times10^{-5}\dimw$ (20\% less) to do the same. 
The compression savings due to the use of prediction are even more apparent for the higher test accuracy which is around $59\%$. 
For this test accuracy $K=1.2\times10^{-4}\dimw$ without prediction and $K=6.5\times10^{-5}\dimw$ with prediction. 
We indicate in the third section of \tbl{topklsummary} how in the latter case the $K$ values translate to rate in bits. 
The number of bits per components is $H_b(K/\dimw) + 32K/\dimw$ as discussed in \secn{numericalwithoutEF}. 
We observe in \tbl{topklsummary} that the use of Est-$K$ predictor offers over a $40\%$ improvement in terms of the bit usage. 

\begin{figure*}
\centering\includegraphics[width=0.9\textwidth]{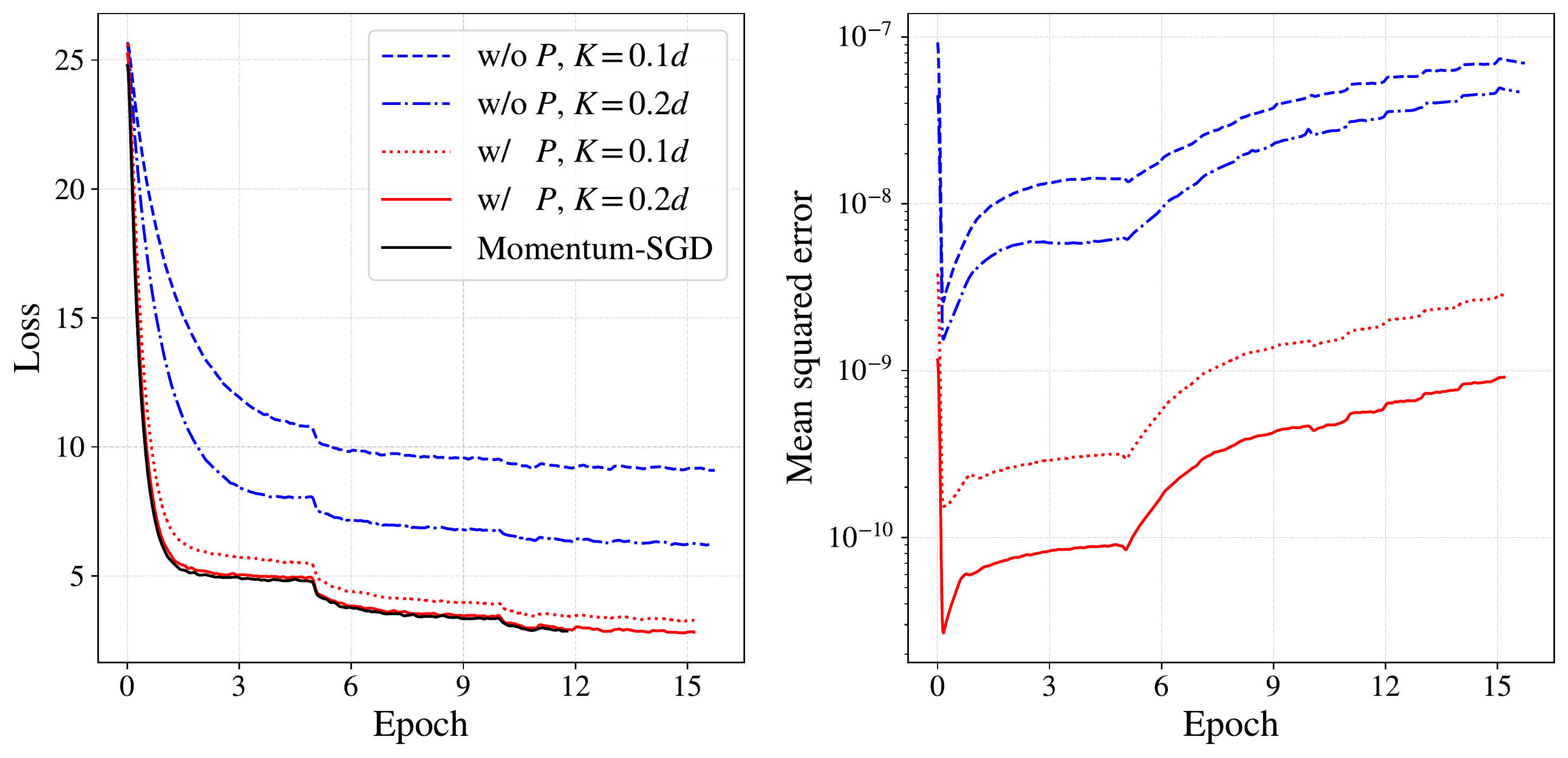}
\caption{Results with the ResNet-50 model with and without the Est-$K$ predictor. The momentum parameter $\beta$ is set to $0.995$. 
The right sub-figure plots $\frac{1}{\dimw}\Lnrm{e_t}^2$.}
\label{fig:predictors_resnet}
\end{figure*}

In \fig{predictors_resnet} we present a comparison of the loss and mean squared error $\frac{1}{\dimw}\Lnrm{e_t}^2$ with and without the Est-$K$ predictor. The results are obtained by training the ResNet-50 model on the original ImageNet dataset (not the downsampled ImageNet-32 version). Additional experimental details are discussed in \secn{numericaliclrtemporal}. We observe that the loss plots of the experiments that employ prediction are much closer to the baseline momentum-SGD plot. In addition, the prediction yields a reduction of the mean squared error close to two orders of magnitude. 

\section{Convergence analysis} \label{secn:convergenceproposed}

In this section we analyze the convergence of $\numDists$-worker
distributed SGD system that employ error-feedback and use
rate-distortion codes for quantization and compression.  Referring
to \fig{block_diagram} we consider systems without momentum
($\beta=0$). Noting that we can write 
\begin{equation}
\ei[i]{t} = \ui[i]{t} - \uti[i]{t} = (\ui[i]{t}+\rhi[i]{t}) - (\uti[i]{t}+\rhi[i]{t}) = \ri[i]{t} - \rti[i]{t}, \label{eqn:eiuirirelationship}
\end{equation}
we simplify the update equations in \eqn{sysModmain} to
\begin{subequations} \label{eqn:equationsSGDEF}
\begin{align}
\ri[i]{t} &= \gi[i]{t} + \frac{\stepsize_{t-1}}{\stepsize_{t}}\ei[i]{t-1} \label{eqn:workerfQinputri} \\
\ei[i]{t} &= \ri[i]{t} - \rti[i]{t} \label{eqn:workererror} \\
\wi{t+1} &= \wi{t} - \stepsize_{t}\avgopn\rti[i]{t}, \label{eqn:workerfQinput}
\end{align}
\end{subequations}
where $\gi[i]{t}$ is the stochastic gradient of $\loss$ at $\wi[i]{t}$
computed by worker-$i$.  
In summary, we show that the convergence rate of this system is asymptotically the same as that of SGD without compression.

Let $\gdi{t}$ be the true gradient of $\loss$
at $\wi{t}$, i.e., $\gdi{t}=\nabla \loss(\wi{t})$.  We make the
following (standard) assumptions.  First, $\loss:\Rel^\dimw\to\Rel$ is
everywhere differentiable with $L$-Lipschitz gradients and $\loss^*
= \min_{\wi{}} \loss(\wi{})$.  Second, $\EV[\gi[i]{t} | \wi{t}]
= \gdi{t}$ and $\EV[\Lnrm{\gi[i]{t}-\gdi{t}}^2] \leq \sigma^2$.  
Let the quantizer $Q$ define a rate distortion code that satisfies $\EV[\Lnrm{\ui[i]{t}-\uti[i]{t}}^2]= \leq D$. 
The expectation is with respect to randomness of $\ui[i]{t}$, the input to the quantizer.  
In other words, the mean squared error of the quantizer is bounded. 
With this assumption, from \eqn{eiuirirelationship} we have 
$\EV[\Lnrm{\ri[i]{t}-\rti[i]{t}}^2]=\EV[\Lnrm{\ei[i]{t}}^2] \leq D$.
\thrm{unusedcompressmain} and \colly{compressedrate} summarize our convergence results for this system. 
Proofs are provided in \apdx{proofsoftheorems}. 

\begin{theorem}\label{thrm:unusedcompressmain}
Let 
$\young>0$, $c=1-\frac{1}{2\young}$ 
and $\stepsize_t=\frac{c}{L\sqrt{T}}$. 
After $T$ iterations, the convergence of the system outlined in \eqn{equationsSGDEF} is characterized by
\begin{align}
\EV\left[\min_{t=0,\dots,T-1}\Lnrm{\nabla \loss(\wi{t})}^2\right]
&\leq \underbrace{\frac{\frac{2L}{c^2}(\loss(\wi{0})-\loss^*) + \frac{\sigma^2}{\numDists} }{2\sqrt{T}-1}}_A
\\
&\qquad + \underbrace{\frac{c\young D}{2{T}-\sqrt{T}}}_B. \label{eqn:stochpropbound}
\end{align}
\end{theorem} 

In comparison to \thrm{unusedcompressmain}, the analogous bound to \eqn{stochpropbound} for SGD without compression is %
\begin{align}
\EV\left[\min_{t=0,\dots,T-1}\Lnrm{\nabla \loss(\wi{t})}^2\right]
&\leq {\frac{2L(\loss(\wi{0})-\loss^*) + \frac{\sigma^2}{\numDists} }{2\sqrt{T}-1}}. \label{eqn:stochsgdbound}
\end{align}
For large $T$, the second term $B$ in~\eqn{stochpropbound} vanishes
faster than the first term $A$.  However, we cannot make a direct
comparison between the first two terms in \eqn{stochpropbound}
and~\eqn{stochsgdbound} when $c^2\neq 1$ in $A$.  Setting an
arbitrarily large value for $\young$ gives $c^2\approx 1$, but we pay
a corresponding penalty via a large $c\young$ in $B$.  The solution
is to make $c$ converge to $1$ asymptotically by carefully choosing a
$\young$ that grows with $T$.  This way $c\to1$ and the first term
in~\eqn{stochpropbound} becomes directly comparable to that
of~\eqn{stochsgdbound}.  We also want $c\young$ in $B$ small enough
so that $B$ vanishes faster than $A$.  We make such a choice of $\young$ in the following \colly{compressedrate}.

\begin{corollary} \label{colly:compressedrate}
Let $\young=T^{1/4}$ in \thrm{unusedcompressmain}. Then we have 
\begin{align}
\EV\left[\min_{t=0,\dots,T-1}\LnrmS{\nabla \loss(\wi{t})}^2\right]
&< \frac{2L(\loss(\wi{0})-\loss^*) + \frac{\sigma^2}{\numDists}}{2\sqrt{T}-1} \\
&\qquad + \frac{2L(\loss(\wi{0})-\loss^*) + D} {2T^{3/4}-T^{1/4}} \\
&\qquad + \bigO\left(\frac{1}{T}\right). \label{eqn:convgrateefsgd}
\end{align}
\end{corollary} 

The second and third terms in \eqn{convgrateefsgd} vanish faster than
the first as $T\to\infty$.  Thus, for large $T$ the system
in \eqn{equationsSGDEF} achieves the same convergence rate
$\bigO(1/\sqrt{T})$ as~\eqn{stochsgdbound} specified for SGD without
compression.

\section{Details of numerical experiments} \label{secn:numericaliclrtemporal}
The details of the experiments presented
in \fig{numericalimagenetcomparecsgd}, \fig{numericalcifcomparecsgd}
and \fig{imagenet32_EFSGD1w} are as follows.  We perform multi-GPU
experiments on a master-worker setup with $4$ workers by training a
wide residual network classifier \cite{zagoruyko2016wide} of depth 28
and width 2 (WRN-28-2).  
The number of trainable parameters (i.e., length of $\dimw$) in this model is close to $1.6$ million. 
Due to the heavy computational needs of
training with the original ImageNet dataset, we use the down-sampled
ImageNet-32 version of ImageNet \cite{chrabaszcz2017downsampled}, in
which images are of size $32\times32$ pixels.  All other properties,
such as the number of images and the number of classes, are the same
as the original ImageNet dataset.  The dataset is partitioned into
four equal sized training sets which are distributed among the
workers.  We implement all algorithms using the TensorFlow and
Horovod \cite{sergeev2018horovod} packages.  Since the communications
component in Horovod is designed for a master-less setup, we simulate
a master-worker environment in our implementation.  In our
experimental setup, a single machine hosts four workers and each
worker runs on a dedicated Tesla P100 16GB GPU.
In all compression algorithms we use blockwise compression, where the gradients corresponding to tensors, matrices and vectors are compressed and decompressed separately. 
Training is started with a learning rate of $0.1$, and the learning rate is multiplied iteratively by $0.1$ at every $8$th epoch. 
We use an $\ellnrm{2}$ weight regularizer scaled by $10^{-4}$. 
Per worker batch size is $64$. 

For the results presented in \fig{predictors_resnet} we train the ResNet-50 model on the original ImageNet dataset (not the downsampled ImageNet-32 version). Per worker batch size is $16$, and the $\ellnrm{2}$ weight regularizer is scaled by $8\times10^{-4}$. Initial learning rate is set to $0.1$, and the learning rate is multiplied iteratively by $0.1$ at every $5$th epoch. All other aspects of the experiments remain same as those with the WRN-28-2 model.

As we summarize in \secn{contrib}, in \fig{imagenet32_compute_time} we present the average computation time per iteration at a worker. This serves as a measure of the additional computational complexity the workers need to complete due to the use of prediction. 
\fig{imagenet32_compute_time} demonstrates that for each quantizer the time per iteration, when using prediction, is only slightly higher than it is when prediction is not used. 
The computational impact to the master node due to its predictor block $P$ can be assessed as follows. 
In iteration $t+1$ the input to $P$ is a vector from iteration $t$. Therefore, in iteration $t+1$ the master can compute the output of the predictor while the workers compute the gradients in iteration $t+1$ and send them to the master. 
As we observe in \fig{imagenet32_compute_time}, the time increment due to the use of $P$ is significantly smaller than that of gradient computations. 
Therefore, it is safe to assume that the delay impact to the master node of computations in $P$ will be minimal. 
We do also note that the proposed Est-$K$ method increases the computational memory requirement. 
However, this effect is also minimal. For example, all experiments we conducted were able to fit even into a low-end GPU, GTX1060-6GB.

\section{Conclusions and future work} \label{secn:conclusions}

In this paper we introduce techniques of gradient quantization and
compression that exploit temporal correlations in consecutive update
vectors to reduce the communication payload.  
We focus on the momentum-SGD algorithm, and we use a predictive coding based approach for the exploitation of temporal correlations. 
In particular, we first consider compression without error-feedback and demonstrate that a simple linear prediction system that exploits the temporal correlations offers significant compression savings. 
Second, we consider compression with error-feedback.
The use of error-feedback allows us to improve the compression ratio
without hindering testing performance. We show that the application of the same linear predictor as before is not useful when error-feedback is employed, therefore we develop a new method we term Est-$K$ that again exploits the temporal correlation. 
Numerical experiments based on the ImageNet dataset show that our algorithms offer significant compression savings compared to schemes that do not make use of temporal correlation.

We design our Est-$K$ predictor to work specifically with the Top-$K$ quantizer. This is because the predictor design becomes simple when the quantizer is Top-$K$, and in the literature Top-$K$ has been shown empirically to work very well when error-feedback is employed. An interesting next step in this theme of works would be to generalize the design ideas of Est-$K$ to accommodate other types of quantizers (such as Scaled-sign and Top-$K$-Q) in systems with error-feedback. 

In the algorithms we derive do not use very advanced temporal prediction. 
For example the predictor we employ for systems with error-feedback implements only a constant estimator (time average of momentum between two updates). 
While our approaches demonstrate that performance improvements result from exploiting temporal correlation, we anticipate that more advanced predictors will perform even better. 
Of course one would need to balance the additional complexity with improved performance. 
One possibility is to optimize the predictor jointly with the quantizer~\cite{joint_design}. 
We leave exploration of advanced predictors to future work.

We also provide a convergence analysis of SGD (without momentum) when compression is applied with error-feedback. 
The literature contains convergence guarantees when the compressors are $\delta$-compressors wherein the compression error is bounded point-wise. 
In our analysis we focus on schemes that provide error bounds only in expectation.
Our analysis can be generalized to include SGD with momentum. 
We leave this as future work.

\section*{Acknowledgment}
The authors would like to thank Jason Lam and Zhenhua Hu of Huawei Technologies Canada for technical discussions, 
and Compute Canada (\href{http://www.computecanada.ca}{www.computecanada.ca}) for providing computing resources for the experiments.

\bibliographystyle{IEEEtran}


\newpage

\appendices

\section{Reasoning for applying momentum at workers} \label{secn:reasoningmastermom}

\begin{figure*}
\centering\includegraphics[width=0.7\textwidth]{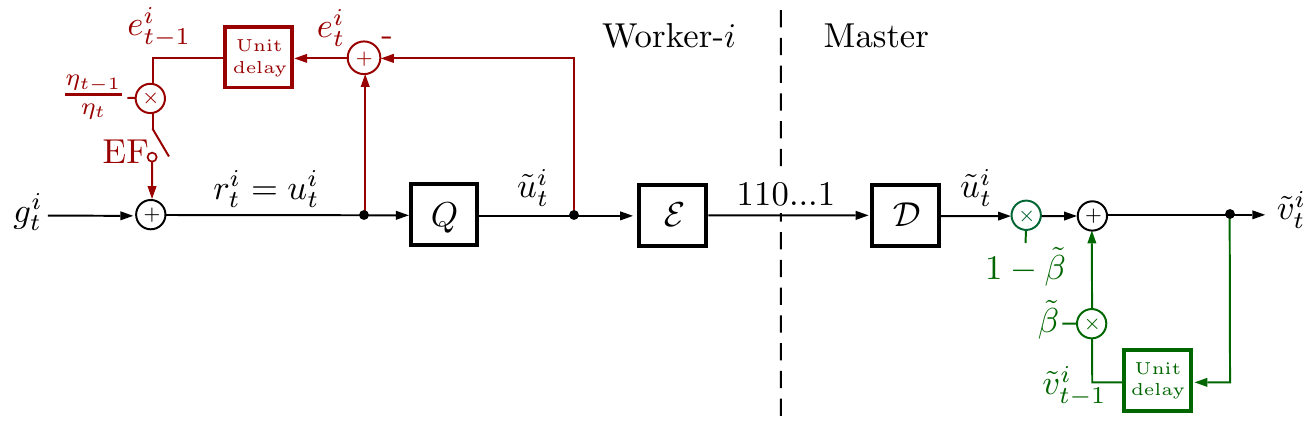}
\caption{System diagram for one worker and master. The master computes the momentum vector.}
\label{fig:block_diagramone}
\end{figure*}

In this section we describe why application of momentum by the master, rather than by each worker, is not desirable. Figure~\ref{fig:block_diagramone} presents the signal flow diagram of a system where worker-$i$ computes the stochastic gradient $\gi[i]{t}$, quantizes it (with the possible application of error-feedback) to obtain $\uti[i]{t}$, and then sends to the master the encoded version of $\uti[i]{t}$. 
Upon receiving $\uti[i]{t}$, the master applies a momentum filter with parameter $\tbeta$ to obtain the momentum vector $\vti[i]{t}$. 
In comparison to \fig{block_diagram}, we do not include the prediction system in \fig{block_diagramone}. This helps to keep the explanation simple. 
Also, in \fig{block_diagramone} we use the notation $\tbeta$ and $\vti[i]{t}$ to differentiate the master-based momentum filter from the worker-based momentum filter considered in \fig{block_diagram} that uses notation $\beta$ and $\vi[i]{t}$. 
We do note that such differentiation of notation is not strictly necessary for the worker-side parameters $\ri[i]{t}$, $\ui[i]{t}$, $\uti[i]{t}$ and $\ei[i]{t}$ because the worker-side of system depicted in \fig{block_diagramone} can be obtained by setting $\beta=0$ in \fig{block_diagram}.

\subsection{Systems without error-feedback}
It is easier first to understand the functioning of the system when the EF switch in \fig{block_diagramone} is open, i.e., without error-feedback. 
Noting that $\uti[i]{k} = \ui[i]{k} - \ei[i]{k} = \gi[i]{k} - \ei[i]{k}$ we can write 
\begin{align}
\vti[i]{t} = \tbeta\vti[i]{t-1} + (1-\tbeta)\uti[i]{t} 
&= (1-\tbeta) \sum_{k=0}^{t} \tbeta^{t-k}\uti[i]{k} \\
&= (1-\tbeta) \sum_{k=0}^{t} \tbeta^{t-k}(\gi[i]{k} - \ei[i]{k}),
\end{align}
which gives 
\begin{equation}
\vti[i]{t}
= (1-\tbeta) \sum_{k=0}^{t} \tbeta^{t-k}\gi[i]{k}
- (1-\tbeta) \sum_{k=0}^{t} \tbeta^{t-k}\ei[i]{k}. \label{eqn:mastermomtwoterms}
\end{equation}

Now, compare the momentum vector with compression in \eqn{mastermomtwoterms} with the momentum vector if there were no quantization error. 
This is equivalent to the worker sending to master $\gi[i]{k}$ without compression. 
We obtain the expression for the momentum without compression by setting $\ei[i]{k}$ in \eqn{mastermomtwoterms} to zero as 
\begin{equation}
\vti[i]{t}
= (1-\tbeta) \sum_{k=0}^{t} \tbeta^{t-k}\gi[i]{k}. \label{eqn:mastermomnocomp}
\end{equation}
Note that we would like to keep the momentum vector with compression (i.e., in \eqn{mastermomtwoterms}) as close as possible to the momentum vector without compression (i.e., in \eqn{mastermomnocomp}). 
Upon comparing the two equations we observe that the second term in \eqn{mastermomtwoterms} causes error to accumulate in the momentum vector. 
This keeps \eqn{mastermomtwoterms} away from the desired value in \eqn{mastermomnocomp}. 
This error accumulation is due to the master's use of a quantized vector $\uti[i]{t}$ (that includes quantization error) in the momentum computation.

\subsection{Systems with error-feedback}
Now, we consider the system in \fig{block_diagramone} when the EF switch is closed, i.e., when error-feedback is used. 
We assume for simplicity that the step size is constant, i.e., $\stepsize_{t-1}=\stepsize_{t}$. By noting that $\uti[i]{k}=\ri[i]{k} - \ei[i]{k}$ and $\ri[i]{k} = \gi[i]{k} + \ei[i]{k-1}$, we get
\begin{align}
\vti[i]{t}
&= (1-\tbeta) \sum_{k=0}^{t} \tbeta^{t-k}\uti[i]{k} \\
&= (1-\tbeta) \sum_{k=0}^{t} \tbeta^{t-k}(\ri[i]{k} - \ei[i]{k}) \\
&= (1-\tbeta) \sum_{k=0}^{t} \tbeta^{t-k}(\gi[i]{k} + \ei[i]{k-1} - \ei[i]{k}) \\
&= (1-\tbeta) \sum_{k=0}^{t} \tbeta^{t-k}\gi[i]{k} + (1-\tbeta) \underbrace{\sum_{k=0}^{t} \tbeta^{t-k}(\ei[i]{k-1} - \ei[i]{k})}_J.
\end{align}
Note that we can rewrite $J$ as
\begin{align}
J
&= \tbeta^{t}\ei[i]{-1} - \ei[i]{t}
+ \sum_{k=1}^{t} \tbeta^{t-k}\ei[i]{k-1} - \sum_{k=0}^{t-1} \tbeta^{t-k}\ei[i]{k} \\
&= - \ei[i]{t} + \sum_{k=0}^{t-1} \tbeta^{t-k-1}\ei[i]{k} - \sum_{k=0}^{t-1} \tbeta^{t-k}\ei[i]{k} \\
&= - \ei[i]{t} + (1-\tbeta)\sum_{k=0}^{t-1} \tbeta^{t-k-1}\ei[i]{k}.
\end{align}
The second equality follows since $\ei[i]{-1}=0$. This gives 
\begin{align}
\vti[i]{t}
&= (1-\tbeta) \sum_{k=0}^{t} \tbeta^{t-k}\gi[i]{k} - (1-\tbeta) \ei[i]{t} \\
&\qquad + (1-\tbeta)^2 \sum_{k=0}^{t-1} \tbeta^{t-k-1}\ei[i]{k}. \label{eqn:mastermomEFcomp}
\end{align}
Again, we would like to keep the result in \eqn{mastermomEFcomp} as close as possible to the momentum vector without compression, i.e., the result in \eqn{mastermomnocomp}. 
The mismatch between \eqn{mastermomnocomp} and \eqn{mastermomEFcomp} stems from the last two terms in \eqn{mastermomEFcomp}. These represent the error accumulation which keeps the momentum vector in \eqn{mastermomEFcomp} away from the desired vector in \eqn{mastermomnocomp}. 
This again demonstrates that computing momentum at the master is not helpful when quantization is used.

\section{Proofs of analytical results} \label{secn:proofsoftheorems}

\subsection{Proof of \thrm{unusedcompressmain}} 
\begin{proof}
Instead of analyzing the sequence $\{\wi{t}\}$, we define a new sequence of vectors $\{\twi{t}\}$ as 
\begin{align}
\twi{t+1} = \wi{t+1} - \stepsize_{t}\avgopn\ei[i]{t}. \label{eqn:newsequence}
\end{align} 
The new sequence satisfies the recurrence relation 
$\twi{t+1} = \twi{t} - \stepsize_{t}\avgopn\gi[i]{t}$ 
which can be verified as follows. 
Substituting to $\wi{t+1}$ and $\ei[i]{t}$ in \eqn{newsequence} from \eqn{workerfQinput} and \eqn{workererror} we obtain 
\begin{align}
\twi{t+1} 
&= \wi{t} - \stepsize_{t}\avgopn\rti[i]{t} - \stepsize_{t}\avgopn(\ri[i]{t} - \rti[i]{t}) \\
&= \wi{t} - \stepsize_{t}\avgopn\ri[i]{t}, 
\end{align}
after which we substitute for $\ri[i]{t}$ from \eqn{workerfQinputri} to get 
\begin{align}
\twi{t+1} 
&= \wi{t} - \stepsize_{t}\avgopn\left(\gi[i]{t} + \frac{\stepsize_{t-1}}{\stepsize_{t}}\ei[i]{t-1}\right) \\
&= \wi{t} - \stepsize_{t-1}\avgopn\ei[i]{t-1} - \stepsize_{t}\avgopn\gi[i]{t} \\
&= \twi{t} - \stepsize_{t}\avgopn\gi[i]{t}. 
\end{align}

Recall that in \secn{convergenceproposed} we define $\gdi{t}$ to be the true gradient of $\loss$
at $\wi{t}$, i.e., $\gdi{t}=\nabla \loss(\wi{t})$. Similarly, let $\gti{t}$ be the gradient of $\loss$ at $\twi{t}$, i.e., $\gti{t}=\nabla \loss(\twi{t})$. From the Lipschitz assumption on $\loss$ we have 
\begin{align}
\loss(\twi{t+1}) 
&\leq \loss(\twi{t}) + \inprd{\gti{t}}{\twi{t+1}-\twi{t}} + \frac{L}{2}\Lnrm{\twi{t+1}-\twi{t}}^2 \\
&= \loss(\twi{t}) - \stepsize_t \inprdS{\gti{t}}{\avgopn\gi[i]{t}} + \frac{L}{2}\stepsize_t^2\LnrmS{\avgopn\gi[i]{t}}^2. \label{eqn:toolongeqn}
\end{align}
Let $H = \EV[\loss(\twi{t+1}) - \loss(\twi{t})]$. 
Taking expectation and rearranging \eqn{toolongeqn} yields 
{\small
\begin{align}
H
&\leq - \stepsize_t \EVS{\inprdS{\gti{t}}{\avgopn\gi[i]{t}}}  + \frac{L}{2}\stepsize_t^2\EVS{\LnrmS{\avgopn\gi[i]{t}}^2} \\
&= - \stepsize_t \underbrace{\EVS{\inprdS{\gti{t}-\gdi{t}}{\avgopn\gi[i]{t}}}}_{H_1}
+ \frac{L}{2}\stepsize_t^2 \underbrace{\EVS{\LnrmS{\avgopn\gi[i]{t}}^2}}_{H_2} \\
&\qquad - \stepsize_t \underbrace{\EVS{\inprdS{\gdi{t}}{\avgopn\gi[i]{t}}}}_{H_3}. \label{eqn:convratedistmain}
\end{align}
}%
Next we bound each term in \eqn{convratedistmain}. 
For $H_2$ in \eqn{convratedistmain} we have 
\begin{align}
H_2 \leq \EV[\Lnrm{\gdi{t}}^2] + \frac{\sigma^2}{\numDists}, \label{eqn:momentbound}
\end{align}
and for $H_3$ in \eqn{convratedistmain} we get 
\begin{align}
H_3
=\EVS{\inprdS{\gdi{t}}{\EVS{\avgopn\gi[i]{t}\middle|\wi{t}}}}
=\EV[\Lnrm{\gdi{t}}^2]. 
\end{align} 
We next bound $-H_1$ (with the negative sign) in \eqn{convratedistmain} as follows. 
First, note that for two given vectors $u,v$ and for any $\young>0$ we have 
$0\leq \Lnrm{\sqrt{\young} u - \frac{1}{\sqrt{\young}}v}^2$, 
which implies $2\langle u,v \rangle \leq \young\Lnrm{u}^2 + \frac{1}{\young}\Lnrm{v}^2$. 
This result is known as the Young's inequality for products (with exponent 2), or as the Peter–Paul inequality. 
For $-H_1$ in \eqn{convratedistmain} we obtain 
\begin{align}
-H_1
&\leq \frac{\young}{2}\EV[\Lnrm{\gti{t}-\gdi{t}}^2] + \frac{1}{2\young}\EVS{\LnrmS{\avgopn\gi[i]{t}}^2} \\
&\leq \frac{\young}{2}L^2\EV[\Lnrm{\twi{t}-\wi{t}}^2] + \frac{1}{2\young}\EV[\Lnrm{\gdi{t}}^2] + \frac{1}{2\young}\frac{\sigma^2}{\numDists} \\
&= \frac{\young}{2}L^2\EVS{\LnrmS{\frac{\stepsize_{t-1}}{\numDists}\sum_{i\in[\numDists]}\ei[i]{t-1}}^2} + \frac{1}{2\young}\EV[\Lnrm{\gdi{t}}^2] + \frac{1}{2\young}\frac{\sigma^2}{\numDists} \\
&\leq \frac{\young}{2}L^2\stepsize_{t-1}^2D + \frac{1}{2\young}\EV[\Lnrm{\gdi{t}}^2] + \frac{1}{2\young}\frac{\sigma^2}{\numDists}. \label{eqn:lastlinefirsttermD}
\end{align}
The first inequality is Young's inequality, the second inequality is due to the Lipschitz assumption on $\loss$ and \eqn{momentbound}, the equality is from \eqn{newsequence}, and the last inequality is obtained as follows. For all $t$ We have 
\begin{align}
\LnrmS{\avgopn\ei[i]{t}}^2
\leq \frac{1}{\numDists^2}\sum_{i,j}\LnrmS{\ei[i]{t}}\LnrmS{\ei[j]{t}}
&= \left(\avgopn\LnrmS{\ei[i]{t}}\right)^2 \\
&\leq \avgopn\LnrmS{\ei[i]{t}}^2. 
\end{align}
The two inequalities are followed by applying Cauchy–Schwarz and Jensen's inequality. 
Taking expectation and asserting the distortion upper bound yields 
\begin{align}
\EVS{\LnrmS{\avgopn\ei[i]{t}}^2}
\leq \avgopn\EV[\LnrmS{\ei[i]{t}}^2]
\leq \avgopn D = D 
\end{align}
which gives \eqn{lastlinefirsttermD}. 

Substituting in all the bounds for the terms on the right hand side of \eqn{convratedistmain} gives 
\begin{align}
H
&\leq \stepsize_t \left(\frac{\young}{2}L^2\stepsize_{t-1}^2D + \frac{1}{2\young}\EV[\Lnrm{\gdi{t}}^2] + \frac{1}{2\young}\frac{\sigma^2}{\numDists}\right) \\
&\qquad - \stepsize_t \EV[\Lnrm{\gdi{t}}^2] 
+ \frac{L}{2}\stepsize_t^2\left(\EV[\Lnrm{\gdi{t}}^2] + \frac{\sigma^2}{\numDists}\right) \\
&= - \stepsize_t \left(1-\frac{L\stepsize_t}{2} - \frac{1}{2\young} \right)\EV[\Lnrm{\gdi{t}}^2] \\
&\qquad + \frac{L}{2}\stepsize_t^2\frac{\sigma^2}{\numDists} 
+ \stepsize_t\stepsize_{t-1}^2 \frac{\young}{2}L^2 D,
\label{eqn:ratedistnoend}
\end{align}
where $H = \EV[\loss(\twi{t+1}) - \loss(\twi{t})]$. 
For now we keep $\young$ as a free variable. Later we will choose the specific value of $\young$ that will result in the tightest bound. Let $c=1-\frac{1}{2\young}$. 
Taking the telescoping sum of \eqn{ratedistnoend} from $t=0$ to $t=T-1$ and rearranging gives 

\begin{align}
\sum_{t=0}^{T-1}\stepsize_t \left(c-\frac{L}{2}\stepsize_t\right)\EV[\Lnrm{\gdi{t}}^2]
&\leq \loss(\wi{0})- \loss^*
+ \frac{L}{2}\frac{\sigma^2}{\numDists}\sum_{t=0}^{T-1} \stepsize_t^2 \\
&\qquad + \frac{\young}{2}L^2D \sum_{t=0}^{T-1}\stepsize_t \stepsize_{t-1}^2, 
\end{align}
where we have asserted that $\twi{0}=\wi{0}$ and that $\EV[\loss(\twi{T})]\geq\loss^*$. 
By noting that the term
\[
\left[\min_{t=0,\dots,T-1}\EV\left[\Lnrm{\gdi{t}}^2\right]\right]\sum_{t=0}^{T-1}\stepsize_t \left(c-\frac{L}{2}\stepsize_t\right)
\]
is less than or equal to 
\[
\sum_{t=0}^{T-1}\stepsize_t \left(c-\frac{L}{2}\stepsize_t\right)\EV[\Lnrm{\gdi{t}}^2],
\]
and noting that 
$\EV[\min_t Z_t] \leq \min_t \EV[Z_t]$ for any finite set of (possibly dependent) random variables $\{Z_t\}$, when $c-\frac{L}{2}\stepsize_t>0$ 
and 
$J = \EV\left[\min_{t=0,\dots,T-1}\Lnrm{\gdi{t}}^2\right]$
we obtain 
\begin{align}
J
&\leq \frac{\loss(\wi{0})-\loss^* 
	+ \frac{\young}{2}L^2D \sum_{t=0}^{T-1}\stepsize_t \stepsize_{t-1}^2 
	+ \frac{L}{2}\frac{\sigma^2}{\numDists}\sum_{t=0}^{T-1} \stepsize_t^2
}{\sum_{t=0}^{T-1} \stepsize_t \left(c-\frac{L}{2}\stepsize_t\right)}. 
\end{align}

Now we are ready to make an appropriate choice for $\stepsize_t$ that satisfies $c-\frac{L}{2}\stepsize_t>0$. 
Note the last two terms in the numerator that include the step sizes raised to the third and second powers, respectively. 
For a diminishing step size we expect the former to vanish more quickly than the latter. 
To this end we pick $\stepsize_t = \frac{c}{L\sqrt{T}}$ to obtain 
\begin{align}
J
&\leq \frac{\loss(\wi{0})-\loss^* 
+ \frac{\young}{2}L^2D \frac{T c^3}{L^3{T}^{3/2}}
+ \frac{L}{2}\frac{\sigma^2}{\numDists}\frac{T c^2}{L^2{T}}
}{T \frac{c}{L\sqrt{T}} \left(c-\frac{c}{2\sqrt{T}}\right)} \\
&= \frac{2L(\loss(\wi{0})-\loss^*) 
+ \young D \frac{c^3}{\sqrt{T}} + c^2\frac{\sigma^2}{\numDists} }{c^2\left(2\sqrt{T}-1\right)}.
\end{align}
Rearranging terms gives \eqn{stochpropbound} which concludes the proof of \thrm{unusedcompressmain}. \qedhere 
\end{proof}

\subsection{Proof of \colly{compressedrate}} \label{secn:ecsgdratecolly}
\begin{proof}
Let $\young=T^s$ for some $s>0$. 
Our choice of $s$ should satisfy two requirements. 
It should make $c\to1$ in the first term $A$ in~\eqn{stochpropbound}, and should make the second term $B\to 0$ faster than $A$. 
We solve for $s$ by expanding $\frac{1}{c^2}$ in~\eqn{stochpropbound}. 
We take the following approach to expand $\frac{1}{c^2}$. 
For $z\in\Rel$ and any positive integer $r$, from the negative binomial series we have 
\[
(1-z)^{-r} = \sum_{p=0}^{\infty} {{r+p-1} \choose p} z^p. 
\]
For $r=2$ and $|z|<1$ we write 
\[
(1-z)^{-2} = 1 + 2z + 3z^2 + 4z^3 + \dots
< 1+2z + \bigO(z^2). 
\]
Since $\frac{1}{2\young} = \frac{1}{2T^s}<1$, we can write 
\begin{align}
\frac{1}{c^2} = 
\left(1-\frac{1}{2\young}\right)^{-2} 
&< 1 + \young^{-1} + \bigO\left(\young^{-2}\right) \\
&= 1 + T^{-s} + \bigO\left(T^{-2s}\right). 
\end{align}
For the term $B$ in~\eqn{stochpropbound} we have $c\young = \young-\frac{1}{2}$. 
Substituting the bound for $\frac{1}{c^2}$ in~\eqn{stochpropbound} and rearranging terms gives 
\begin{align}
J
&< \frac{2L(\loss(\wi{0})-\loss^*)  + \frac{\sigma^2}{\numDists} }{2\sqrt{T}-1} \\
&\qquad + \left(T^{-s} + \bigO\left(T^{-2s}\right)\right)
\frac{2L(\loss(\wi{0})-\loss^*) }{2\sqrt{T}-1} \\
&\qquad + \left(T^{s}-\frac{1}{2}\right)\frac{D}{2{T}-\sqrt{T}}  \label{eqn:beforebigo} \\
&< \frac{2L(\loss(\wi{0})-\loss^*)  + \frac{\sigma^2}{\numDists} }{2\sqrt{T}-1} \\
&\qquad + \bigO\left(T^{-s-\frac{1}{2}}\right) + \bigO\left(T^{-2s-\frac{1}{2}}\right) \\
&\qquad + \bigO\left(T^{s-1}\right) + \bigO\left(T^{-1}\right),
\end{align}
where $J = \EV\left[\min_{t=0,\dots,T-1}\Lnrm{\gdi{t}}^2\right]$.
We would like to pick an $s$ that makes all terms in $\bigO$ notation go to zero 
at the same asymptotic rate. 
The $\bigO\left(T^{s-1}\right)$ term requires $s<1$. 
For $0<s<1$, $\bigO\left(T^{s-1}\right)$ and $\bigO\left(T^{-s-\frac{1}{2}}\right)$ converge to zero most slowly and therefore they determine $s$. 
Setting $s-1 = -s-\frac{1}{2}$ gives $s=\frac{1}{4}$. 
Substituting this choice for $s$ in \eqn{beforebigo} gives 
\begin{align}
J
&< \frac{2L(\loss(\wi{0})-\loss^*) + \frac{\sigma^2}{\numDists} }{2\sqrt{T}-1} + \bigO\left(\frac{1}{T}\right) \\
&\qquad + \frac{2L(\loss(\wi{0})-\loss^*) + D}{2T^{3/4}-T^{1/4}}
- \frac{\frac{1}{2}D}{2{T}-\sqrt{T}}. 
\end{align}
Combining the second two terms in the two lines yields \eqn{convgrateefsgd}, concluding the proof of \colly{compressedrate}. 
\qedhere
\end{proof}

\end{document}